\documentclass[review]{elsarticle}

\usepackage{hyperref}

\usepackage{amsfonts}
\usepackage{amsmath}
\usepackage{graphicx,setspace,algorithmic,algorithm,multirow,cases,amssymb,subfigure,setspace}
\usepackage{lettrine}
\usepackage{multirow}
\usepackage{xcolor}
\usepackage{times}
\usepackage{graphicx}
\usepackage{amsmath}
\usepackage{indentfirst}
\usepackage{booktabs}
\usepackage{blindtext}
\usepackage{array}
\usepackage{tabularx,booktabs}
\usepackage{color, soul}
\usepackage{bm}
\usepackage{array}
\usepackage{booktabs}

\biboptions{sort&compress}

\journal{Journal of Knowledge-Based Systems}

\begin{document}

\begin{frontmatter}
\title{Audio-Visual Collaborative Representation Learning for Dynamic Saliency Prediction}

\author[a]{Hailong Ning}
\author[b]{Bin Zhao\corref{d}}
\cortext[d]{Corresponding author}
\author[a]{Zhanxuan Hu}
\author[a]{Lang He}
\author[a]{Ercheng~Pei}

\address[a]{School of Computer Science and Technology, Xi’an University of Posts and Telecommunications, Xi’an 710121, P. R. China.}
\address[b]{School of Artificial Intelligence, Optics and Electronics (iOPEN), Northwestern Polytechnical University, Xi'an 710072, P. R. China.}

\begin{abstract}
The Dynamic Saliency Prediction (DSP) task simulates the human selective attention mechanism to perceive the dynamic scene, which is significant and imperative in many vision tasks. Most of existing methods only consider visual cues, while neglect the accompanied audio information, which can provide complementary information for the scene understanding. In fact, there exists a strong relation between auditory and visual cues, and humans generally perceive the surrounding scene by collaboratively sensing these cues. Motivated by this, an audio-visual collaborative representation learning method is proposed for the DSP task, which explores the audio modality to better predict the dynamic saliency map by assisting vision modality. The proposed method consists of three parts: 1) audio-visual encoding, 2) audio-visual location, and 3) collaborative integration parts. Firstly, a refined SoundNet architecture is adopted to encode audio modality for obtaining corresponding features, and a modified 3D ResNet-50 architecture is employed to learn visual features, containing both spatial location and temporal motion information. Secondly, an audio-visual location part is devised to locate the sound source in the visual scene by learning the correspondence between audio-visual information. Thirdly, a collaborative integration part is devised to adaptively aggregate audio-visual information and center-bias prior to generate the final saliency map. Extensive experiments are conducted on six challenging audiovisual eye-tracking datasets, including DIEM, AVAD, Coutrot1, Coutrot2, SumMe, and ETMD, which shows significant superiority over state-of-the-art DSP models.
\end{abstract}


\end{frontmatter}


\section{Introduction}
Saliency prediction task aims to automatically predict the most prominent area in the scene by simulating the human selective attention mechanism, which provides an alternative for obtaining the most valuable information from massive data. The task has served an important research topic in the field of computer vision, and can be of great applications in many fields, such as scene understanding \cite{lai2022weakly, jian2021integrating}, object detection \cite{fang2022lc3net, ji2022dmra}, object tracking \cite{Zhang2017Online}, image quality evaluation \cite{zhu2021saliency}, automatic contrast enhancement \cite{yang2020saliency}, and video compression \cite{zhu2018spatiotemporal}.

In the field of computer vision, the saliency prediction task draws increasing attention, and lots of methods have been proposed in recent years \cite{itti1998model, bruce2007attention, gao2009decision, hou2007saliency, wang2017deep}. According to the data type, the existing methods can be divided into two aspects, including Static Saliency Prediction (SSP) methods and Dynamic Saliency Prediction (DSP) methods. The SSP methods aim to leverage the low-level contrast information and high-level semantic information of images to achieve the prediction of prominent area in the scene \cite{cong2018review}. As an early exploration of the saliency prediction task, Itti {\it et al.} \cite{itti1998model} imitate the human bottom-up visual selective attention process to extract the low-level features of images, predict the corresponding saliency map.
With the popularity of deep learning \cite{ma2022multi, cheng2022tsgb, wang2022hierarchical}, a large number of researchers conduct the saliency prediction task by mining high-level semantic information in images \cite{mopuri2018cnn, xu2018personalized, wang2021semantic, wang2017deep}. For example, Wang {\it et al.} \cite{wang2017deep} obtain the hierarchically saliency information by extracting multi-scale features of images. The DSP methods focus on applying the spatio-temporal structure information in the video for the prediction of prominent area in the scene. For example, Zhang {\it et al.} \cite{zhang2020spatial} design a spatial-temporal DSP model for learning spatial features with a static network and temporal features with a dynamic network.
In view of practical applications, this paper aims to investigate the saliency prediction for the dynamic video.


\subsection{Motivation and Overview}\label{motivation}
Generally, video data naturally includes two modalities, {\it i.e.}, audio and vision. They represent the scene content in the video from different aspects, and can complement to help the viewer better understand the video content. Recently, lots of multi-modal studies based on audio-visual data have shown that audios can significantly promote the understanding of the scene \cite{di2020Discriminative, di2020cross, song2022multimodal, qian2020multiple}. For example, Hu {\it et al.} \cite{di2020cross} proposed a cross-task transfer learning model for scene classification based on audio-visual data, which shows the benefits of audio-visual analysis compared with single-modality analysis.

However, most researchers in DSP have not fully realized the potential contribution of audio information to the performance. They predict the dynamic saliency map by only mining the information in the visual modality itself, while ignoring the latent effect of the accompanied audio information.

It is well known that human attention is naturally influenced by audio-visual stimuli rather than only auditory or visual stimuli in isolation. Inspired by this, we argue that audio information can assist the vision modality to better predict the saliency map in this paper.
Specifically, humans practically pay more attention to the sounding object in the video. With the assistance of audio-visual information, it is easily to locate more salient sounding object based on audio-visual consistency learning. In addition, human instinctively pays more attention to some other cues, such as the moving, center, and high-level semantic objects.
However, there are often some inconsistencies among the influences for DSP by multiple cues. For example, 1) when existing multiple moving objects in a scene, how to locate the more prominent sounding objects; 2) When the original audio is replaced with background sound, or when the sounding objects is not in the filed of the view, how to locate the salient objects; 3) As for the effect by multiple cues, {\it i.e.} movement, sound, and center-prior, how to better integrate them. These inconsistencies will bring some interference and are the main challenges for the audio-visual DSP.

Based on the above opportunities and challenges, an audio-visual collaborative representation learning method is proposed for the DSP task. The proposed method can be purposely leveraged to predict the dynamic saliency map in videos by collaboratively integrating the audio-visual cues. Concretely, the proposed method is composed of three main parts: 1) audio-visual encoding, 2) audio-visual location, and 3) collaborative integration parts. Firstly, the audio-visual encoding part consists of two branches, which are responsible for learning audio and visual features with spatial location and temporal motion information, respectively. Specifically, the audio branch adopts a refined SoundNet architecture \cite{aytar2016soundnet}, and the visual branch employs a modified 3D ResNet-50 architecture \cite{hara2018can}. Secondly, with the proposed audio-visual location part, the learned audio-visual features are jointly to locate the sound source in the visual scene by learning the correspondence between audio-visual information. Thirdly, the collaborative integration part is introduced to adaptively aggregate multi-cues audio-visual information, so as to generate the final dynamic saliency map. Here, the multi-cues information indicates the localized sound source information, spatio-temporal visual information (motion, high-level semantics, {\it etc.}), as well as center-bias prior.

\subsection{Contributions}\label{contribution}
Generally, the main contributions of this paper are threefold:
\begin{itemize}
  \item An audio-visual collaborative representation learning method is proposed, which can comprehensively consider the influence of audio-visual cues on dynamic saliency prediction, and realize the role of auxiliary enhancement.
  \item An audio-visual location part is devised to learn the correspondence between audio and visual modalities, so as to predict the sounding objects in the scene and reduce interference.
  \item A collaborative integration part is designed, which can adaptively aggregate the influence of multiple cues on dynamic saliency prediction.
\end{itemize}

\subsection{Organization}\label{organization}
The remaining sections of the paper are arranged as follows: In Section \ref{relatedworks}, the related works are reviewed. In Section \ref{TheProposedmethod}, we describe the proposed method in details. In Section \ref{experiments}, the experimental results are shown and discussed. And in Section \ref{conclutions}, the conclusion is drawn.

\begin{figure*}
\begin{center}
\includegraphics[width=\linewidth]{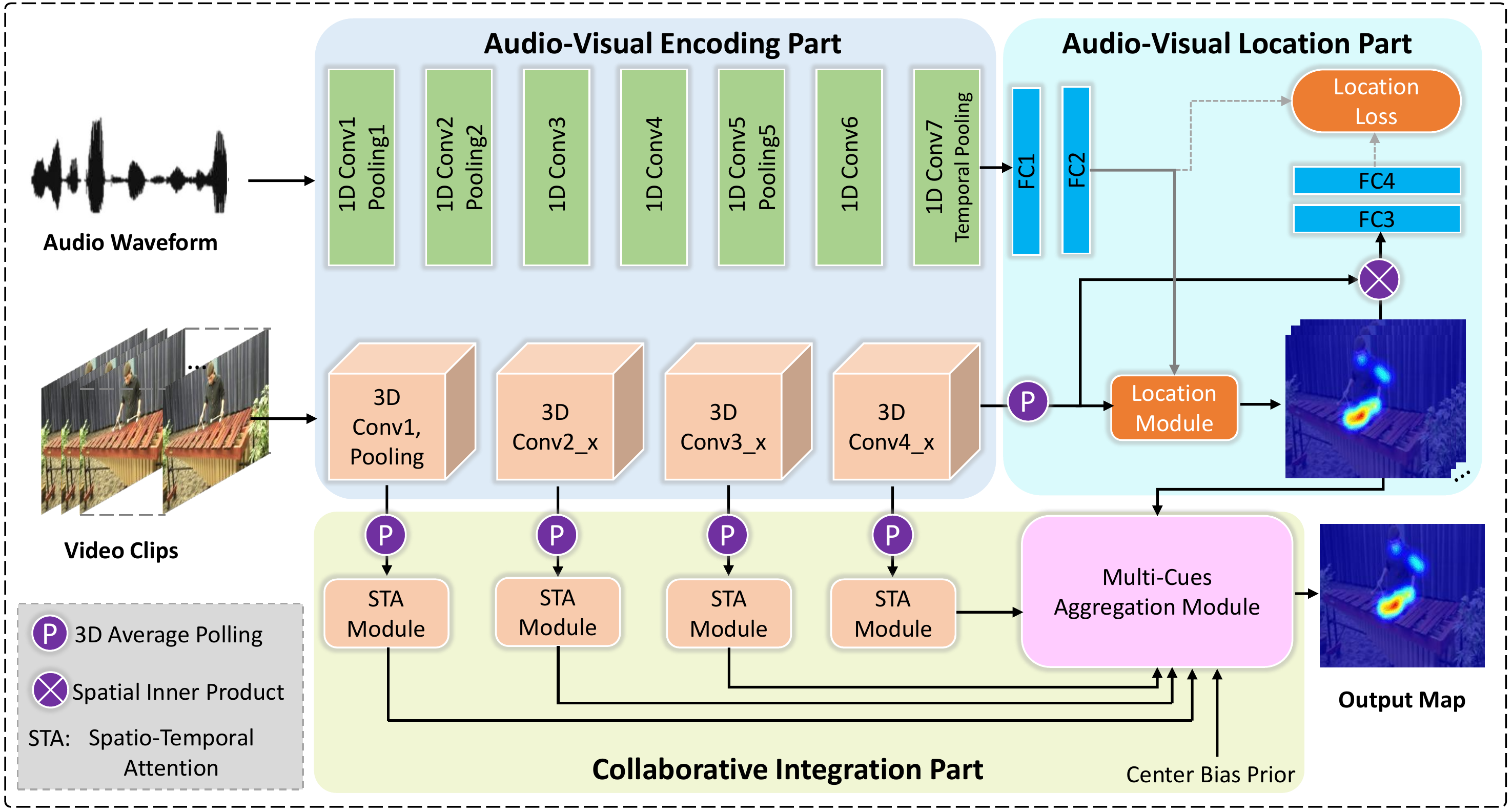}
\renewcommand{\figurename}{Fig.}
\vspace{-8 mm}
\end{center}
    \caption{\small{The framework of the proposed method. Firstly, an audio-visual encoding part is designed to learn audio-visual features. Secondly, an audio-visual location part is devised to locate the sound source in the visual scene by learning the correspondence between audio-visual information. Thirdly, a collaborative integration part is proposed to adaptively aggregate audio-visual information and center-bias prior to generate the final dynamic saliency map.}}
\label{fig:1}
\end{figure*}

\section{Related works}\label{relatedworks}

In recent years, numerous saliency prediction methods have been proposed. According to the data type, the existing methods can be divided into two aspects, including Static Saliency Prediction (SSP) methods and Dynamic Saliency Prediction (DSP) methods. This section reviews the existing saliency prediction methods in the following successively.

\subsection{Static Saliency Prediction Methods}\label{image}
Earlier works devote to investigate the saliency prediction task based on static images \cite{liu2021visual, wang2018attention, zhang2021saliency, kroner2020contextual}. These works are mostly on the basis of bottom-up visual attention mechanism \cite{liu2021visual, zhang2021uncertainty, judd2009learning}.
Itti {\it et al.} \cite{itti1998model} firstly conducted the saliency prediction task by imitating the human bottom-up visual selective attention process to extract the low-level visual features of images. To match actual eye movements, Judd {\it et al.} \cite{judd2009learning} considered both bottom-up visual attention and top-down image semantics, and collect a large eye tracking dataset to address the problem.
On the assumption of contrast, many computational methods on SSP have been proposed. Perazzi {\it et al.} \cite{perazzi2012saliency} designed an intuitive SSP method based on the contrast feature. Wang {\it et al.} \cite{wang2012visual} leveraged selective contrast, including color, texture, and location, to predict the salient regions of images.
In addition, many researchers have realized the SSP from the aspects of information-theoretic \cite{bruce2007attention}, decision-theoretic \cite{gao2009decision}, and spectral analysis \cite{hou2007saliency}, {\it etc.}. Since the adopted features in these methods are hand-crafted, large-scale data with complex distributions cannot be well processed.

As a result, most researchers begin to conduct the SSP task by mining high-level semantic information in images with deep neural networks \cite{mopuri2018cnn, xu2018personalized, wang2017deep}. Vig {\it et al.} \cite{Vig2014Large} firstly proposed utilizing deep neural networks for the SSP task. However, due to the insufficient training data, the performance is limited. To address this problem, Jiang {\it et al.} \cite{Jiang2015SALICON} built the SALICON dataset with plenty of natural images and the corresponding eye-tracking data. Based on the dataset, lots of works on SSP are developed by the follow-up researchers \cite{he2019understanding}. He {\it et al.} \cite{he2019understanding} explored the intrinsical reason of the large gap between deep models and the inter-human baseline. In addition, more effective network architectures are exploited for learning representative features \cite{Wang2018Deep, Sss2017DeepFix, Liu2018Learning, wang2018attention, zhang2021saliency, kroner2020contextual}. Zhang {\it et al.} \cite{zhang2021saliency} proposed to incorporate prior knowledge of semantic relationships so as to learn highlighted regions in images. Kroner {\it et al.} \cite{kroner2020contextual} developed an encoder-decoder structure to learn multi-scale high-level visual features for SSP. Considering that both low-level contrast features and high-level semantic features are important for SSP, Yuan {\it et al.} \cite{yuan2019bio} introduced a bio-inspired representation learning method to generate the saliency map. Wang {\it et al.} \cite{Wang2018Deep} conducted the SSP by fusing features from multiple layers of VGG-16.
Kruthiventi {\it et al.} \cite{Kummerer2017Understanding} considered the centre-bias prior information and developed a computational method for SSP, which improves the predicted results considerably.

\subsection{Dynamic Saliency Prediction Methods}\label{video}
With the tremendous progress of data storage technology and mobile Internet technology, massive video data are soaring recently \cite{li2017multiview, liVideoDistillation2021}. In order to deal with the vast amounts of information, researchers pay more attention on DSP to predict the most valuable information in video data \cite{jiang2018deepvs, bak2017spatio, gorji2018going, sun2018sg, lai2019video, wu2020salsac, wang2020video, chen2021video}. Jiang {\it et al.} \cite{jiang2018deepvs} developed an object-to-motion convolutional neural network for predicting the intra-frame visual saliency. Bak {\it et al.} \cite{bak2017spatio} presented a spatio-temporal saliency network for DSP. Gorji {\it et al.} \cite{gorji2018going} proposed a multi-stream ConvLSTM structure with the attentional push effect of the scene actors and the photographer. To improve the performance of DSP models, Sun {\it et al.} \cite{sun2018sg} put forward a step-gained fully convolutional network by simultaneously considering motion and temporal information.
By conducting multi-scale feature learning and spatio-temporal feature integration, Lai {\it et al.} \cite{lai2019video} designed a residual attentive learning network for DSP. Wu {\it et al.} \cite{wu2020salsac} presented an end-to-end neural network named SalSAC for DSP. The network is based on CNN-LSTM-Attention and integrates both static and dynamic information. In order to fully consider the effect of both global and local consistency on DSP, Wang {\it et al.} \cite{wang2020video} introduced a dynamic saliency network on the basis of both global discriminations and local consistency. These methods have greatly promoted the progress of DSP. Nevertheless, the effect of the audio information accompanying the video is ignored when conducting the DSP task.

Considering the influence of audio cues on the vision task, a few attempts have been made to better perceive the scene information and predict the saliency map. Starting from application-specific, some researcher adopt the traditional signal processing techniques for locating the salient region in the scene \cite{min2016fixation, sidaty2017toward, min2020multimodal}. For example, Min {\it et al.} \cite{min2020multimodal} utilized cross-modal kernel canonical correlation analysis to predict the moving-sounding object. Subsequently, more and more attention is paid on salient regions location by integration of audio \cite{zhu2021lavs, qian2020multiple, hu2020discriminative, Afouras20b, tsiami2020stavis, chen2021audiovisual}. Qian {\it et al.} \cite{qian2020multiple} proposed a two-stage audio-visual learning method for visually localizing multiple sound sources in unconstrained videos. To locate sounding objects in cocktail-party, Hu {\it et al.} \cite{hu2020discriminative} introduced a two-stage learning framework with a self-supervised class-aware manner. Afouras {\it et al.} \cite{Afouras20b} develop a model using attention to transform a video into several discrete audio-visual objects. Tavakoli {\it et al.} \cite{tavakoli2019dave} designed a conceptually simple and effective audio-visual analysis method for dynamic saliency prediction.
Tsiami {\it et al.} \cite{tsiami2020stavis} proposed a spatio-temporal audio-visual saliency network by combining both visual and
auditory information.

This paper is dedicated to conduct the dynamic saliency prediction task by exploring the collaborative mechanism among different audio-visual cues, while alleviating existing inconsistencies between audio-visual modalities. 

\section{The proposed network}\label{TheProposedmethod}
In this work, an audio-visual collaborative representation learning method is proposed to generate dynamic saliency map. As is shown in Figure \ref{fig:1}, the proposed method is composed of three main parts: 1) audio-visual encoding, 2) audio-visual location, and 3) collaborative integration parts. Thereinto, the audio-visual encoding part is responsible for learning audio-visual features, which contain spatial location and temporal motion information. The audio-visual location part aims at locating the sound source in the visual scene by learning the correspondence between audio-visual features. The collaborative integration part is in charge of adaptively aggregating the localized sound source information, spatio-temporal visual information (motion, high-level semantics, {\it etc.}), and center-bias prior information, so as to generate the final dynamic saliency map.
These parts are expatiated successively as follows.

\subsection{Audio-Visual Encoding}\label{featureExtraction}
The audio-visual encoding part consists of two paralleled branches for encoding the original data as audio semantic features and spatio-temporal visual features, respectively. Details about the two branches are elaborated in the following subsections.

\subsubsection{\textbf{Audio Encoding}}\label{Audio_encoding}
In the DSP task based on audio-visual analysis, it is important to obtain the semantic concept of audio rather than low-level signal \cite{senocak2018learning}. To this end, the audio signals are represented by using convolutional neural networks (CNNs). Specifically, we follow the previous work \cite{tsiami2020stavis} and adopt the 1-D fully convolutional network for processing the audio waveform. Firstly, the audio segment is cropped to match the visual frames duration ({\it i.e.} 16 frames). Secondly, a Hanning window is leveraged to acquire the central audio value with a higher weight, which represents the current time instance, and model the past and future attenuation values. Thirdly, a 1-D fully convolutional network with the first seven layers of the SoundNet \cite{aytar2016soundnet} and a temporal max-pooling layer is applied for encoding high-level semantic features. Formulaically, the process of audio encoding can be written as:
\begin{equation}\label{Eq.:1}
{\bf{S}}_A=\mathcal F_{A}({\bf{X}}_A;\theta_{A}),
\end{equation}
where ${\bf{X}}_A$ and ${\bf{S}}_A$ represent the input audio data and the corresponding high-level semantic feature, respectively. $\mathcal F_{A}$ stands for the mapping function from audio data to the corresponding high-level semantic feature. $\theta_{A}$ is the parameter during the process of audio encoding.

\subsubsection{\textbf{Spatio-Temporal Visual Encoding}}\label{Visual_encoding}
In order to capture the spatial semantic information and temporal motion information, the 3-D CNNs is adopted for processing the video frames. Specifically, the 3D ResNet-50 architecture \cite{hara2018can}, which is proposed initially for action recognition, is employed as the backbone to encode the spatio-temporal features. Lots of works \cite{chen2020deep, pang2020multi, zhang2019hierarchical} have demonstrated that multi-scale features contribute to achieving a good performance for perceiving objects with different scales. As a result, the multi-scale features are introduced for DSP in this work. As is shown in Figure \ref{fig:1}, the spatio-temporal visual encoding branch adopts the first 4 ResNet convolutional blocks to provide the outputs ${\bf S}_V^{1}, {\bf S}_V^{2}, {\bf S}_V^{3}, {\bf S}_V^{4}$, which contain different spatial and temporal information. The process of spatio-temporal visual encoding can be written as:

\begin{equation}\label{Eq.:2}
{\bf{S}}_V^m=\mathcal F_{V}^m({\bf{X}}_V;\theta_{V}^{m}),
\end{equation}
where ${\bf{X}}_V$ and ${\bf{S}}_V^m$ stand for the input video frames and the corresponding spatio-temporal feature of the $m$-th ResNet convolutional block. $\mathcal F_{V}^m$ is the mapping function from video frames to the corresponding spatio-temporal feature. $\theta_{V}^m$ is the parameter during the process of spatio-temporal visual encoding.

\subsection{Audio-Visual Location}
In this subsection, an audio-visual location part is devised to locate the sounding object by exploiting the consistency between the audio and visual modalities in a sharing latent space. By this way, the related sounding objects are selectively and dynamically brought out to the foreground when audio and visual concepts appear simultaneously. For example, the playing the piano, barking dog, {\it etc.}, in the video, are expected to be detected after the audio-visual location.

The audio-visual location can be implemented with 4 steps. Firstly, the output of the 4-th ResNet convolutional block is selected as spatio-temporal visual feature, termed as ${\bf S}_V^{4}$, since it contains rich semantic information for the visual frames. In order to marginal out the temporal dimension and acquire a global representation, a temporal average pooling operation is applied for ${\bf S}_V^{4}$. For simplicity, the global representation is denoted as a reshaped matrix form ${\bf V}=\left[{\bf v}_1; \cdots; {\bf v}_B\right] \in \mathbb{R}^{B \times D_{h}}$. Secondly, to match the dimension of the global representation in visual branch and higher level concept of audio signal, two fully connected (FC) layers with ReLU activation are applied for the audio feature ${\bf{S}}_A$, so as to generate the audio embedding ${\bf h}_{A}$ with $D_{h}$-dimension. Thirdly, an attention mechanism is utilized in the location module (see Figure \ref{alg:Framwork}) to locate the sounding object and generate a audio-aware saliency map. It is achieved as follows:
\begin{equation}\label{Eq.:3}
a_{b}=\left<{\bf W}_{1}{\bf v}_{b}, {\bf W}_{2}{\bf h}_{A}\right>,
\end{equation}

\begin{equation}\label{Eq.:4}
\alpha_{b} = \frac{{\rm exp}(a_{b})}{\sum_{b=1}^{B}{{\rm exp}(a_{b}})},
\end{equation}
where $a_{b}$ captures the dependency between ${\bf h}_{A}$ and ${\bf v}_{b}$. ${\bf W}_{1}$ and ${\bf W}_{2}$ are the training weights. $\left<\cdot,\cdot \right>$ means the inner-product operation between two matrixes. $\alpha_{b}$ stands for the $b$-th element in the sounding map ${\bm \alpha}$, which can be interpreted as the probability of location related to the audio context. Further, the audio-aware saliency map ${\bf F}_{\rm audio}$ can be computed by a upsampling operation. Note that the upsampling operation adopts the resize-convolution operation rather than deconvolution to avert the checkerboard effect\footnote{https://distill.pub/2016/deconv-checkerboard/}.
After obtaining the sounding map, we fourthly compute the representative context vector ${\bf h}_{z}$ to interact the sounding map with the global representation in visual branch at the sound source location. The process is achieved by:

\begin{equation}\label{Eq.:5}
{\bf h}_{z} = \sum_{b=1}^{B}{{\alpha}_{b}{\bf v}_{b}}.
\end{equation}

Then, the representative context vector ${\bf h}_{z}$ is transformed to a spatio-temporal visual feature $\hat{\bf v}$ with two fully connected (FC) layers with ReLU activation. Finally, a location loss is imposed on $\hat{\bf v}$ and ${\bf{S}}_A$ to learn the features to share latent space of the audio and visual modality. Here, the Euclidean distance is employed to calculate the location loss.

\subsection{Collaborative Integration}
As mentioned before, human attention is influenced by many aspects, including the sounding, moving, center, high-level semantic objects, {\it etc.}. To model these aspects, this subsection presents a collaborative integration mechanism. Specifically, a spatio-temporal attention module is proposed to detect the moving and semantic objects. A learnable center-bias prior function is introduced to generate center-bias prior-aware map. A multi-cues aggregation module is devised to integrate the influence of different cues and generate the final saliency map.


\subsubsection{\textbf{Moving and High-Level Semantic Objects Prediction}}\label{motion}
In order to model the influence of moving and semantic objects for DSP, a spatio-temporal attention module is proposed. The spatio-temporal attention module contains two branches for capturing the moving information and semantic information, respectively.

As for the temporal attention branch, the temporally moving information is modeled. As is shown in Figure \ref{Eq.:2} (a), the spatio-temporal visual feature from $m$-th ResNet block is firstly processed with the operation of averaging pooling in the channel dimension, resulting in ${{\bf S}_{V,T}^{m}}$. Secondly, the first and last frames are removed, respectively, leading to ${{\bf S}_{V,T,-1}^{m}}$ and ${{\bf S}_{V,T,0}^{m}}$. Thirdly, the temporal attention is computed by conducting the frame-wise similarity between ${{\bf S}_{V,T,-1}^{m}}$ and ${{\bf S}_{V,T,0}^{m}}$, so as to capture the temporally moving information. Specifically, the frame-wise similarity can be implemented as:
\begin{equation}\label{Eq.:6}
{\bf M}_{T}^{m}=\sum_{t=1}^{t=T-1}{\left(1-\left({\bf S}_{V,t,0}^{m}-{\bf S}_{V,t,-1}^{m}\right)\right)},
\end{equation}
where ${\bf M}_{T}^{m}$ represent the motion feature from the m-th ResNet convolutional block. $t$ indexes the frames. Afterwards, the motion feature is further processed with a $1\times1$ convolution and a resize-convolution operation to generate a motion-aware saliency map ${\bf F}_{\rm motion}^{m}$.

\begin{figure}
\begin{center}
\includegraphics[width=\linewidth]{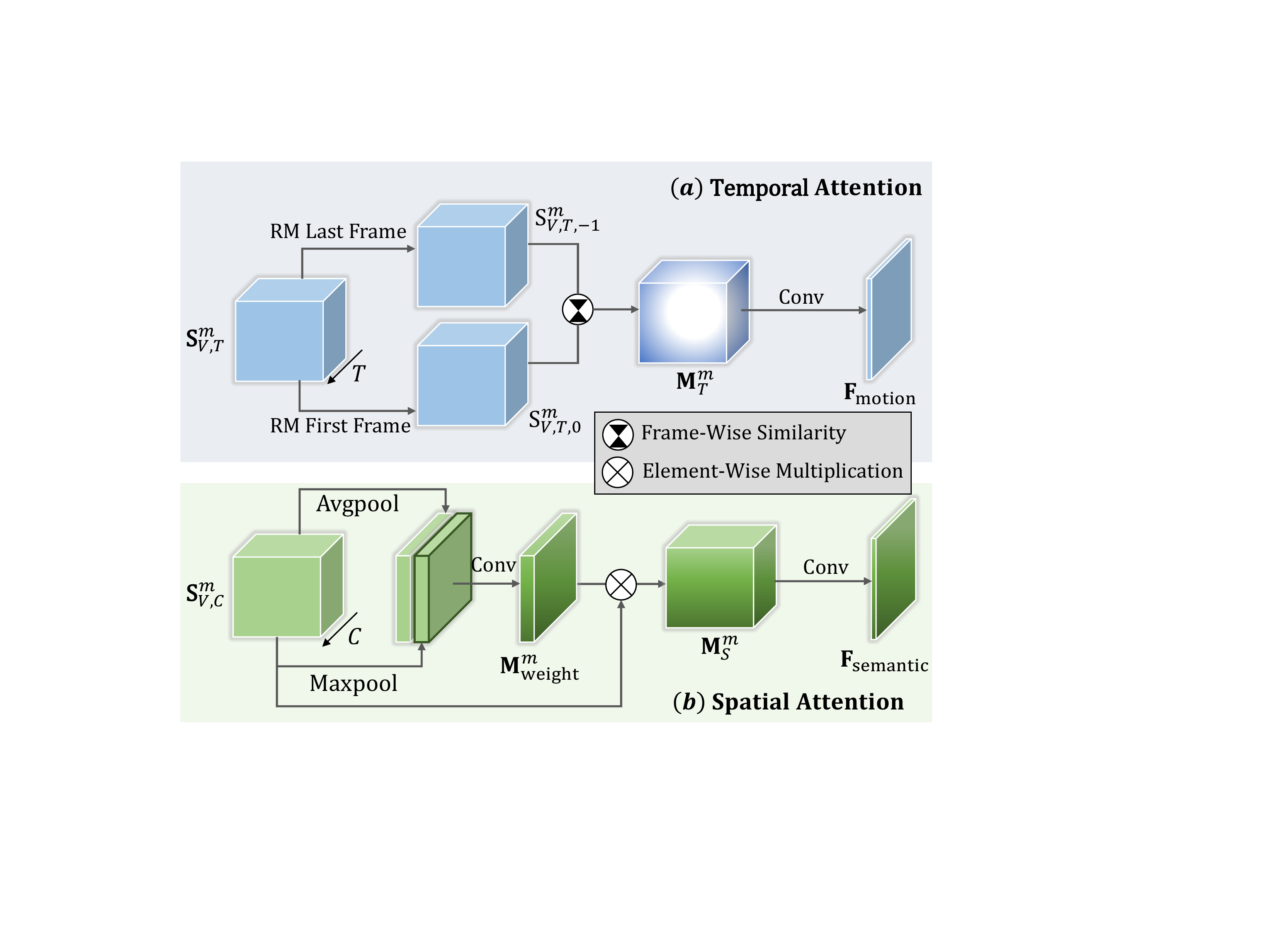}
\renewcommand{\figurename}{Fig.}
\vspace{-8 mm}
\end{center}
    \caption{\small{The devised spatio-temporal attention module.}}
\label{fig:2}
\end{figure}

In respect to the spatial attention branch, the high-level semantic information is expected to be captured by leveraging the inter-spatial relationship of features. As is shown in Figure \ref{fig:2} (b), the spatio-temporal visual feature from $m$-th ResNet block is firstly processed with the operation of averaging pooling in the temporal dimension, so as to obtain ${\bf S}_{V,C}^{m}$. Secondly, we apply the max-pooling and average-pooling along with the channel dimension, and concatenate them as an efficient feature descriptor. In this way, the highlighting informative regions can be effectively shown \cite{komodakis2017paying, woo2018cbam}. Thirdly, a convolution layer with  filter size of $7\times7$ is conducted for generating a spatial weight matrix ${\bf M}_{\rm weight}^{m}$. Fourthly, by employing an element-wise multiplication between ${\bf S}_{V,C}^{m}$ and the spatial weight matrix ${\bf M}_{\rm weight}^{m}$, the feature ${\bf M}_{S}^{m}$ with high-level semantic information is learned. Finally, ${\bf M}_{S}^{m}$ is further processed with the $1\times1$ convolution and resize-convolution operation to generate a high-level semantic-aware saliency map ${\bf F}_{\rm semantic}^{m}$.

\subsubsection{\textbf{Center-Bias Prior}}\label{center}
According to previous studies \cite{cornia2018predicting, droste2020unified, zhang2021saliency}, human attention tends to concentrate on the center of scenes, which is termed as center-bias phenomenon. To this end, a learnable center-bias prior function is adopted according to our preceding work \cite{yuan2019bio}. Specifically, the center-bias prior-aware map ${\bf{F}}_{\rm center}$ is generated using a Gaussian function as follows:

\begin{equation}\label{Eq.:7}
{\bf{F}}_{\rm center}=\frac{1}{2\pi\sigma_{x}\sigma_{y}}{\rm exp}\left(-\left(\frac{(x-x_{0})^2}{2\sigma^2_{x}}+\frac{(y-y_{0})^2}{2\sigma^2_{y}}\right)\right),
\end{equation}
where $\sigma^2_{x}$ and $\sigma^2_{y}$ indicate the to-be-learned horizontal variance and vertical variance, respectively. The generated center-bias prior-aware map ${\bf{F}}_{\rm center}$ represents a spatial pattern. Note that the center-bias information is modeled purely from learning.

\subsubsection{\textbf{Multi-Cues Aggregation}}\label{fuse}

We have computed the audio-aware saliency map, the motion-aware saliency map, the high-level semantic-aware saliency map, and the center-bias prior-aware saliency map. They can express the saliency driven by different cues. As a result, it is quite essential to integrate them for generating the final saliency map. For this purpose, a multi-cues aggregation module is proposed for integrating the influence of different cues, by exploiting the consistency among them and reducing the difference.

\begin{figure}
\begin{center}
\includegraphics[width=\linewidth]{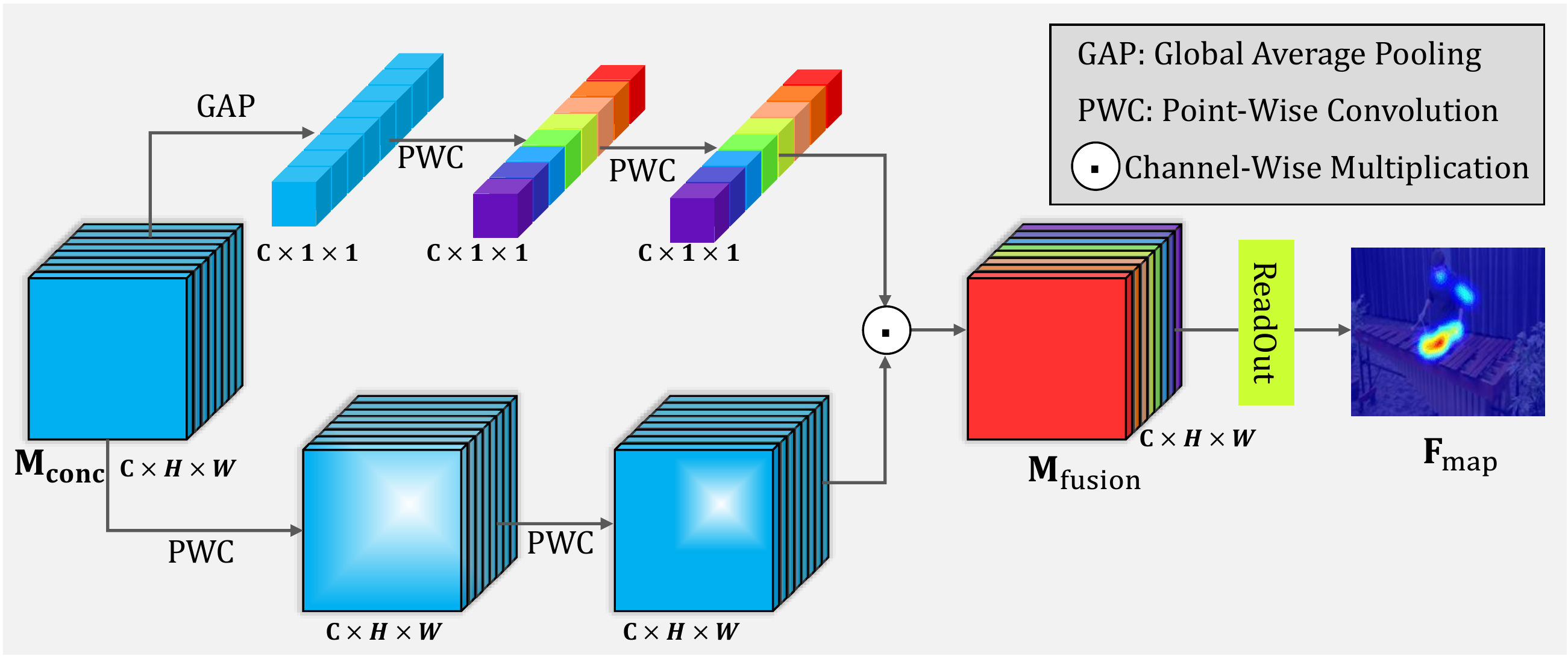}
\renewcommand{\figurename}{Fig.}
\vspace{-8 mm}
\end{center}
    \caption{\small{The proposed multi-cues aggregation module.}}
\label{fig:3}
\end{figure}

As is shown in Figure \ref{fig:3}, the multi-cues aggregation is conducted by two branches. One branch is to learn the global channel context, and the other branch is responsible for perceiving local channel context. The outputs of the two branches are combined to obtain the fused multi-cues feature. Specifically, the multi-cues aggregation module is composed of three main steps. Firstly, the concatenated feature ${\bf{M}}_{\rm conc}=\left[{\bf{F}}_{\rm audio}; {\bf{F}}_{\rm motion}; {\bf{F}}_{\rm semantic}; {\bf{F}}_{\rm center}\right]$
is processed as the global response context ${\bf{g}}$ by:

\begin{equation}\label{Eq.:8}
\begin{split}
{\bf{g}}&={\rm global}({\bf{M}}_{\rm conc};{\bf{W}}_{3})\\
&={\sigma}\left({\mathcal B}\left(\rm{PWC}\left({\sigma}\left({\mathcal B}\left(\rm{PWC}\left(\rm{GAP}({\bf{M}}_{\rm conc})\right)\right)\right)\right)\right)\right),
\end{split}
\end{equation}
where $\rm global$ denotes the global response mapping function. ${\bf{W}}_{3}$ is the to-be-learned parameter. $\sigma$ represents the Sigmoid function. ${\mathcal B}$ means the Batch Normolization (BN) operation. $\rm{PWC}$ stands for the Point-Wise Convolution (PWC), which is chosen for its lightweight. $\rm{GAP}$ represents the global average pooing.

In parallel, the local response context ${\bf{L}}$ can be computed by:
\begin{equation}\label{Eq.:9}
\begin{split}
{\bf{L}}&={\rm local}({\bf{M}}_{\rm conc};{\bf{W}}_{4})\\
&={\mathcal B}\left(\rm{PWC}\left({\delta}\left({\mathcal B}\left(\rm{PWC}\left({\bf{M}}_{\rm conc}\right)\right)\right)\right)\right),
\end{split}
\end{equation}
where $\rm local$ denotes the local response mapping function. ${\bf{W}}_{4}$ is the to-be-learned parameter. $\delta$ represents the ReLU function.

Secondly, to better integrate the influence of different cues on DSP, the global response context and local response context are combined as:
\begin{equation}\label{Eq.:10}
{\bf{M}}_{\rm fusion}={\bf{g}} \odot {\bf{L}},
\end{equation}
where ${\bf{M}}_{\rm fusion}$ represents the fused feature by considering different cues, and $\odot$ means the channel-wise multiplication.

Finally, based on the fused feature ${\bf{M}}_{\rm fusion}$, the final saliency map ${\bf{F}}_{\rm map}$ is computed by a Readout Network, which is composed of three successive $1\times1$ convolution layers.

\subsection{Optimizing Strategy}
In order to obtain the final saliency map, we aggregate the saliency map driven by cues of sounding, moving, center, high-level semantic objects. Let ${\bf{Y}}_{sal}$ represent the ground-truth fixation map obtained by the eye-tracking data. During audio-visual location process, the objective function is defined as:
\begin{equation}\label{Eq.:11}
\begin{split}
\mathcal{L}_{\rm A}&=\rm{KL}\left({\bf{F}}_{\rm audio},{\bf{Y}}_{\rm sal}\right)+\|(\hat{\bf{v}},{\bf{S}}^{A})\|_{2}\\
&={\bf{Y}}_{\rm sal}\log\left(\frac{{\bf{Y}}_{\rm sal}}{{\bf{F}}_{\rm audio}+\epsilon}+\epsilon\right)+\|(\hat{\bf{v}},{\bf{S}}^{A})\|_{2},
\end{split}
\end{equation}
where $\rm{KL}(\cdot,\cdot)$ stands for the Kullback–Leibler divergence between the two distributions. $\|\cdot,\cdot\|_{2}$ is the Euclidean distance. $\epsilon$ indicates the regularization constant to avoid the NaN value in the loss.

During the moving and high-level semantic objects prediction process, the objective function is defined as:
\begin{equation}\label{Eq.:12}
\begin{split}
\mathcal{L}_{\rm MS}=&\rm{KL}\left({\bf{F}}_{\rm motion},{\bf{Y}}_{\rm sal}\right)+\rm{KL}\left({\bf{F}}_{\rm semantic},{\bf{Y}}_{\rm sal}\right)\\
=&{\bf{Y}}_{\rm sal}\log\left(\frac{{\bf{Y}}_{\rm sal}}{{\bf{F}}_{\rm motion}+\epsilon}+\epsilon\right)\\
&+{\bf{Y}}_{\rm sal}\log\left(\frac{{\bf{Y}}_{\rm sal}}{{\bf{F}}_{\rm semantic}+\epsilon}+\epsilon\right).
\end{split}
\end{equation}

During the multi-cues aggregation process, the objective function is defined as:
\begin{equation}\label{Eq.:13}
\begin{split}
\mathcal{L}_{\rm fuse}&=\rm{KL}\left({\bf{F}}_{\rm map},{\bf{Y}}_{\rm sal}\right)\\
&={\bf{Y}}_{\rm sal}\log\left(\frac{{\bf{Y}}_{\rm sal}}{{\bf{F}}_{\rm map}+\epsilon}+\epsilon\right).
\end{split}
\end{equation}

Ultimately, the final loss $\mathcal{L}_{\rm final}$ of training all parameters can be jointly combined by the losses $\mathcal{L}_{A}$, $\mathcal{L}_{MS}$, and $\mathcal{L}_{fuse}$, as follows.
\begin{equation}\label{Eq.:14}
\mathcal{L}_{\rm final}=w_{1}\mathcal{L}_{\rm A}+w_{2}\mathcal{L}_{\rm MS}+w_{3}\mathcal{L}_{\rm fuse},
\end{equation}
where $w_{1}$, $w_{2}$, and $w_{3}$ are the tradeoff coefficients controlling the contribution of each term.

\begin{algorithm}[htb]
\renewcommand{\algorithmicrequire}{\textbf{Input:}}
\renewcommand\algorithmicensure {\textbf{Output:} }
\caption{The proposed method}
\label{alg:Framwork}
\begin{algorithmic}[1]
\REQUIRE ~~\\
Training video frames ${\bf{X}}_{V}$ and the corresponding audio data ${\bf{X}}_{A}$;\\
Testing video frames ${\bf{X}}_{V}^{te}$ and the corresponding audio data ${\bf{X}}_{A}^{te}$.
\ENSURE ~~\\
Testing saliency map ${\bf{F}}_{\rm map}^{te}$;\\
All the to-be-learned parameters $\bf{W}$.
\renewcommand{\algorithmicrequire}{\textbf{Initialization:}}
\REQUIRE ~~\\
The parameter $\theta_{A}$ in the audio encoding branch is initialized from the origin SoundNet \cite{aytar2016soundnet}.
The parameter $\theta_{V}$ in the spatio-temporal visual encoding branch is initialized from the origin 3D ResNet-50 \cite{hara2018can}.
The remaining weights are randomly initialized by truncated\_normal distribution.
\renewcommand{\algorithmicrequire}{\textbf{Repeat:}}
\REQUIRE ~~\\
\STATE Calculate the high-level semantic feature ${\bf{S}}_{A}$ and spatio-temporal feature ${\bf{S}}_{V}$ according to Eq. \ref{Eq.:1} and Eq. \ref{Eq.:2}, respectively;
\STATE Generate the audio-aware saliency map ${\bf{F}}_{\rm audio}$ according to Eq. \ref{Eq.:3} and Eq. \ref{Eq.:4};
\STATE Calculate the motion-aware saliency map ${\bf{F}}_{\rm motion}$, the high-level semantic-aware saliency map ${\bf{F}}_{\rm semantic}$, and center-bias prior-aware map ${\bf{F}}_{\rm center}$ based on Section \ref{motion} and Section \ref{center};
\STATE Generate the final saliency map ${\bf{F}}_{\rm map}$ based on Section \ref{fuse};
\STATE Compute the final loss $\mathcal{L}_{\rm final}$ according to Eq. \ref{Eq.:14};
\STATE Update all the parameters by utilizing Adam optimizer.
\renewcommand\algorithmicensure {\textbf{Until:}{\,}{A fixed number of iterations.}}
\ENSURE ~~\\
\STATE Generate the testing saliency map ${\bf{F}}^{te}_{\rm map}$.
\renewcommand\algorithmicensure {\textbf{Return:}{\,}{${\bf{F}}^{te}_{\rm map}$, $\bf{W}$}}
\ENSURE ~~\\
\end{algorithmic}
\end{algorithm}

Based on the final loss $\mathcal{L}_{\rm final}$, the proposed method can be optimized as follows. The parameter $\theta_{A}$ in the audio encoding part are initialized from the origin SoundNet \cite{aytar2016soundnet}. The parameter $\theta_{V}$ in the spatio-temporal visual encoding part is initialized from the origin 3D ResNet-50 \cite{hara2018can}, which is pretrained on the Kinetics 400 dataset for action recognition task. The remaining weights are randomly initialized by truncated\_normal distribution. In the training stage, the optimizing process is composed of five main steps. Firstly, the training video frames ${\bf{X}}_{V}$ and the corresponding audio data ${\bf{X}}_{A}$ are processed as high-level semantic feature ${\bf{S}}_{A}$ and spatio-temporal feature ${\bf{S}}_{V}$ with the audio-visual encoding part. Secondly, the high-level semantic feature ${\bf{S}}_{A}$ and spatio-temporal feature ${\bf{S}}_{V}$ are combined to locate the sounding object to generate the audio-aware saliency map ${\bf{F}}_{\rm audio}$ by exploiting the consistency in a sharing latent space of the audio and visual modality. Thirdly, the motion-aware saliency map ${\bf{F}}_{\rm motion}$, the high-level semantic-aware saliency map ${\bf{F}}_{\rm semantic}$, and center-bias prior-aware map ${\bf{F}}_{\rm center}$ are computed based on Section \ref{motion} and Section \ref{center}; Fourthly, the final saliency map ${\bf{F}}_{\rm map}$ are inferred based on ${\bf{F}}_{\rm audio}$, ${\bf{F}}_{\rm motion}$, ${\bf{F}}_{\rm semantic}$, and ${\bf{F}}_{\rm center}$. Finally, we compute the final loss $\mathcal{L}_{\rm final}$ based on the generated final saliency map ${\bf{F}}_{\rm map}$ and the ground-truth fixation map ${\bf{Y}}_{\rm sal}$, and update all the parameters $\bf{W}$ by minimizing $\mathcal{L}_{\rm final}$ with Adam optimizer. Once the training epoch reaches 50, the training process is terminated. Afterwards, the parameter $\bf{W}$ is utilized to infer the testing saliency map ${\bf{F}}^{te}_{\rm map}$. It is to note that the proposed method is trained in an end-to-end manner. The details about the optimization process are shown in Algorithm \ref{alg:Framwork}.

\section{Experiment and results}\label{experiments}
The experiments are conducted on six benchmark datasets with audio-visual eye-tracking data.
In the following subsections, the implementation details, evaluation metrics are elaborated. In addition, the experimental results are given and analyzed from the aspects of ablation study and comparison with the state-of-the-arts.

\subsection{Setup}

\subsubsection{\textbf{Datasets}}
The proposed method is trained and evaluated on AVAD \cite{min2016fixation}, Coutrot1 \cite{coutrot2016multimodal, coutrot2014saliency}, Coutrot2 \cite{coutrot2016multimodal, coutrot2014saliency}, DIEM \cite{mital2011clustering}, ETMD \cite{koutras2015perceptually, tsiami2019behaviorally} and SumMe \cite{gygli2014creating, tsiami2019behaviorally} datasets. These datasets contains various types videos accompanied with audios. Specifically, 1) the AVAD dataset \cite{min2016fixation} contains 45 video clips with each duration 5-10 seconds. It covers various audio-visual activities, {\it e.g.}, playing the piano, playing basketball, making an interview, {\it etc.}. The dataset also contains the eye-tracking data from 16 participants. 2) The Coutrot1 and Coutrot2 datasets are split from the Coutrot dataset \cite{coutrot2016multimodal, coutrot2014saliency}. The Coutrot1 dataset is with 60 video clips covering 4 visual categories: one moving object, several moving objects, landscapes, and faces. The corresponding eye-tracking data are recorded from 72 participants. The Coutrot2 dataset includes 15 video clips, which record 4 persons having a meeting. The corresponding eye-tracking data are from 40 participants. 3) The DIEM dataset \cite{mital2011clustering} involves 84 video clips, including game trailers, music videos, advertisements, {\it etc.}. Note that the audio and visual tracks do not correspond naturally. The eye-tracking data are recorded via 42
participants. The ETMD dataset \cite{koutras2015perceptually, tsiami2019behaviorally} includes 12 video clips from several hollywood movies. The eye-tracking data are annoted by 10 different people. The SumMe dataset \cite{gygli2014creating, tsiami2019behaviorally} consists of 25 video clips with diverse topics, {\it e.g.}, playing ball, cooking, traveling, {\it etc.}. The corresponding eye-tracking data are collected from 10 viewers.

Following the previous work \cite{tsiami2020stavis}, we adopt the same data partitioning for traning and testing. Specifically, 3 different splits of the data are created with non-overlapping among train, validation and test sets. The performance are evaluated by taking the average among all 3 splits.

\subsubsection{\textbf{Evaluation Metrics}} In order to measure the consistency between the predicted saliency map and the groundtruth fixation map, 5 widely-used evaluation metrics for DSP are employed \cite{bylinskii2018different}. The evaluation metrics include CC, NSS, AUC-Judd (AUC-J), shuffled AUC (sAUC), and SIM. The CC measures the linear correlation coefficient between the groundtruth fixation map and the predicted saliency map. The NSS aims at measuring the saliency value on human fixations. The AUC-J and sAUC are location-based metrics for evaluate the predicted saliency map. The SIM measures the similarity between the predicted saliency map and groundtruth fixation map. The 5 evaluation metrics provide a comprehensive assessment for DSP.

\subsubsection{\textbf{Implementation Details}} The input samples are processed as 16 video frames and the corresponding audio stream. Each video frame is resized at $112\times112$ pixels. Following the previous work \cite{tsiami2020stavis}, the data augmentation is also employed for random generation of training samples. The implementation adopts the 3D ResNet-50 \cite{hara2018can} as backbone for encoding spatio-temporal visual features, and applies SoundNet \cite{aytar2016soundnet} as backbone for encoding high-level audio semantic features. The Gaussian kernel for generating center-bias prior map is with size of $7\times7$. The tradeoff coefficients $w_{1}$, $w_{2}$, and $w_{3}$ in Eq. \ref{Eq.:14} are all selected as 1 empirically. The batchsize is set as 128. The proposed method is optimized by utilizing the Adam optimizer with learning rate of $10^{-4}$.
When the iterative epoch reaches 50, the optimizing process is terminated. During the test process, a sliding window method is adopted for
inferring the final saliency map of each frame. The experiment is conducted by the $Pytorch$ library and on the PC with a TITAN RTX GPU and 24G RAM.

\subsection{Ablation Analysis}
In this subsection, we aims at verifying the effectiveness of several main parts for the proposed method. Specifically, six different variations are constructed, including:
\begin{itemize}
\item {\textbf{Visual Model:}} Only visual information is leveraged for DSP, while the audio information is ignored.
\item {\textbf{AV Inner-Product:}} The audio-visual location is implemented by directly conducting inner-product operation, rather than by adopting the proposed audio-visual location part to exploit the consistency between the audio and visual modalities in a sharing latent space.
\item {\textbf{Concatenate Fusion:}} In order to integrate the multi-cues maps and generate the final saliency map, we directly concatenate them and further readout, rather than by utilizing the proposed multi-cues aggregation module.
\item {\textbf{Proposed (w/o SA):}} The spatial attention is not considered for modeling higher-level semantic information.
\item {\textbf{Proposed (w/o TA):}} The temporal attention is not considered for modeling motion information.
\item {\textbf{Proposed:}} The complete method proposed by us.
\end{itemize}

TABLE \ref{tab:1} exhibits the results of different variations. By the observation and analysis from the results, we can verify five main observations:

\begin{table*}
\caption{The ablation study on six benchmark datasets.}
\centering
\scriptsize
\setlength{\tabcolsep}{1.5mm}{
\begin{tabular}{c|ccccc|ccccc}
\hline
\multirow{2}{*}{Methods}&
\multicolumn{5}{c|}{AVAD}&
\multicolumn{5}{c}{Coutrot1} \\
\cline{2-11}
  &CC$\uparrow$	   &NSS$\uparrow$	   &AUC-J$\uparrow$   &sAUC$\uparrow$   &SIM$\uparrow$    &CC$\uparrow$	   &NSS$\uparrow$	   &AUC-J$\uparrow$   &sAUC$\uparrow$   &SIM$\uparrow$ \\
\hline
Visual Model        &0.5714     &3.11	&0.8928	&0.5740	    &0.4635	&0.4458	&2.03	&0.8527	&0.5562	&0.3630	 \\
AV Inner-Product	&0.5751     &3.14	&0.9004	&0.5827	    &0.4572	&0.4608	&2.16	&0.8604	&0.5681	&0.3744	 \\
Concatenate Fusion	&0.5826     &3.32	&0.9081	&0.5944	    &0.4680	&0.4714	&2.29	&0.8702	&0.5774	&0.3853	 \\
Proposed (w/o SA)	&0.5907     &3.36	&0.9127	&0.6014	    &0.4752	&0.4807	&2.35	&0.8696	&0.5842	&0.3967	 \\
Proposed (w/o TA)	&0.6125     &3.44	&0.9204	&0.6145	    &0.4783	&0.4852	&2.36	&0.8704	&0.5861	&0.4017	 \\
\bf{Proposed} &\bf{0.6262} &\bf{3.57}	&\bf{0.9251}	&\bf{0.6203}	    &\bf{0.4820}	&\bf{0.4985}	&\bf{2.44}	&\bf{0.8798}	&\bf{0.6042}	&\bf{0.4154}	 \\
\hline\hline
\multirow{2}{*}{Methods}&
\multicolumn{5}{c}{Coutrot2}&
\multicolumn{5}{c}{DIEM}\\
\cline{2-11}
   &CC$\uparrow$	   &NSS$\uparrow$	   &AUC-J$\uparrow$   &sAUC$\uparrow$   &SIM$\uparrow$    &CC$\uparrow$	   &NSS$\uparrow$	   &AUC-J$\uparrow$   &sAUC$\uparrow$   &SIM$\uparrow$ \\
\hline
Visual Model	    &0.4260	&3.26	&0.9208	&0.6344	&0.3071	&0.5461     &2.08	&0.8544	&0.6351     &0.4335 \\
AV Inner-Product	&0.6748	&4.68	&0.9370	&0.6727	&0.4283	&0.5542     &2.15	&0.8716	&0.6552	    &0.4551 \\
Concatenate Fusion	&0.7034	&4.91	&0.9435	&0.6911	&0.4824 &0.5763     &2.21	&0.8835	&0.6740	    &0.4673 \\
Proposed (w/o SA)	&0.7221	&5.09	&0.9466	&0.7028	&0.4960	&0.5816     &2.25	&0.8847	&0.6781	    &0.4726 \\
Proposed (w/o TA)	&0.7283	&5.14	&0.9473	&0.7135	&0.5048	&0.5834     &2.29	&0.8903	&0.6844	    &0.4830 \\
\bf{Proposed} &\bf{0.7481}	&\bf{5.45}	&\bf{0.9537}	&\bf{0.7294}	&\bf{0.5266}	&\bf{0.5924}     &\bf{2.33}	&\bf{0.8941}	&\bf{0.6982}    &\bf{0.4917} \\
\hline

\hline\hline
\multirow{2}{*}{Methods}&
\multicolumn{5}{c|}{ETMD}&
\multicolumn{5}{c}{SumMe} \\
\cline{2-11}
   &CC$\uparrow$	   &NSS$\uparrow$	   &AUC-J$\uparrow$   &sAUC$\uparrow$   &SIM$\uparrow$    &CC$\uparrow$	   &NSS$\uparrow$	   &AUC-J$\uparrow$   &sAUC$\uparrow$   &SIM$\uparrow$ \\
\hline
Visual Model	    	&0.4817	&2.41	&0.9180	&0.6842	&0.3504	&0.3482	&1.86	&0.8573	&0.6235	&0.2840 \\
AV Inner-Product	    &0.5024	&2.53	&0.9247	&0.6981	&0.3584	&0.3615	&1.92	&0.8650	&0.6361	&0.3015 \\
Concatenate Fusion	    &0.5346	&2.81	&0.9283	&0.7217	&0.4011	&0.4037	&2.16	&0.8742	&0.6527	&0.3281 \\
Proposed (w/o SA)	    &0.5284	&2.73	&0.9217	&0.7185	&0.3952	&0.4164	&2.18	&0.8780	&0.6613	&0.3327 \\
Proposed (w/o TA)		&0.5415	&2.99	&0.9274	&0.7261	&0.4133	&0.4228	&2.21	&0.8807	&0.6653	&0.3371 \\
\bf{Proposed}        &\bf{0.5664}	&\bf{3.05}	&\bf{0.9351}	&\bf{0.7406}	&\bf{0.4325}	&\bf{0.4392}	&\bf{2.25}	&\bf{0.8945}	&\bf{0.6712}	&\bf{0.3428} \\
\hline
\end{tabular}}
\label{tab:1}%
\end{table*}

\begin{figure*}
\begin{center}
\includegraphics[width=\linewidth]{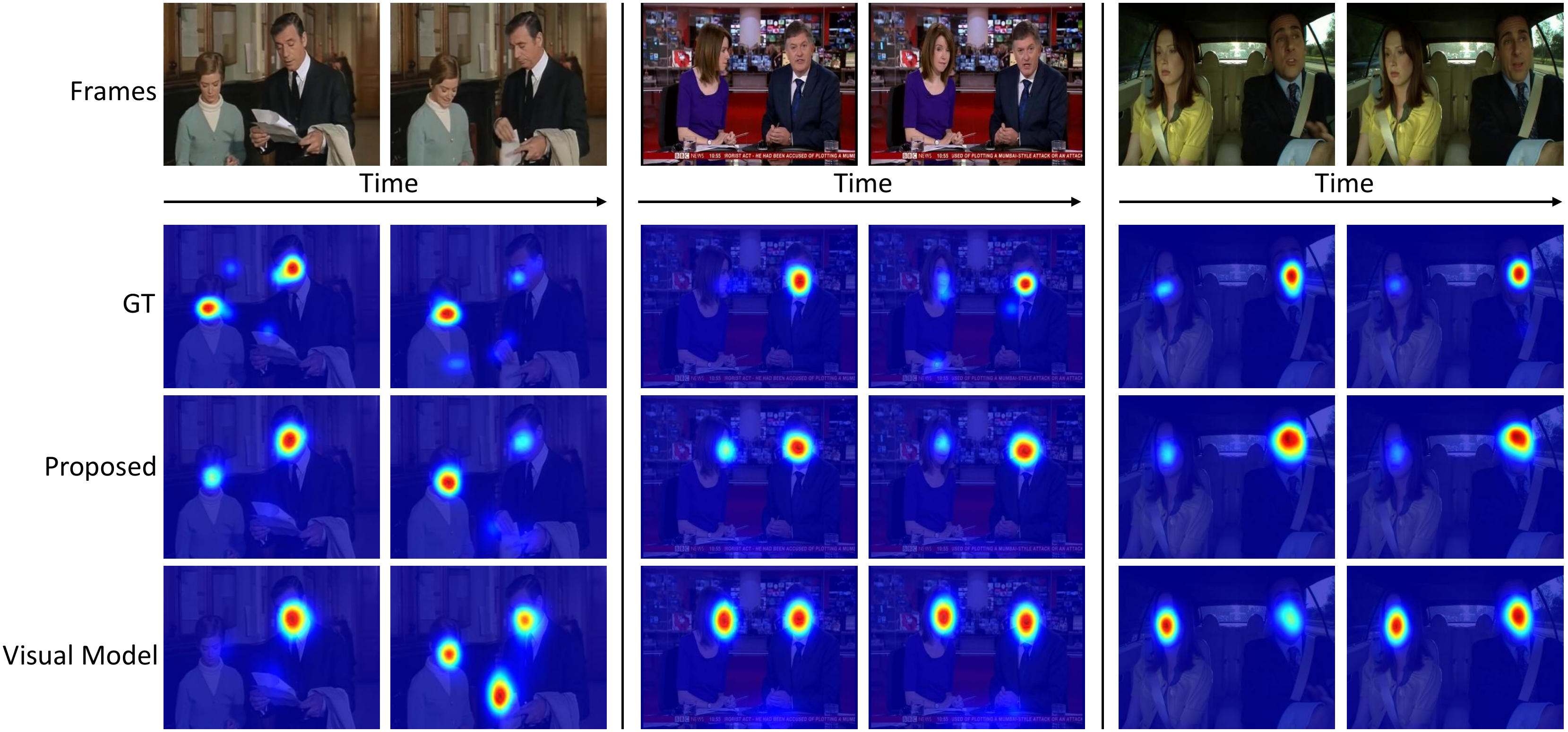}
\renewcommand{\figurename}{Fig.}
\vspace{-5 mm}
\end{center}
    \caption{\small{Some examples by adopting different settings. The first row shows the raw frames. The second row shows the corresponding groundtruth (GT) of the saliency map. The third row exhibits the predicted saliency maps when audio and visual information is employed simultaneously. The last row displays the predicted saliency maps when only visual information is leveraged.}}
\label{fig:4}
\end{figure*}

1) The audio information plays a significant role on DSP. The conclusion can be supported by the comparison results between Visual Model and the proposed method. Specifically, It is worth noticing that the performance drops significantly when only visual information is leveraged. Concretely, the CC value drops by more than 5\% on the AVAD dataset. The NSS value is dropped form 5.45 to 3.26 on the Coutrot2 dataset. The AUC-J value is decreased from 0.8941 to 0.8544 on the DIEM dataset. The sAUC value falls almost 6\% on the ETMD dataset. And the SIM metric drops nearly 8\% on the SumMe dataset. The results demonstrate the important role of audio information for the DSP task. Especially, the results on the Coutrot2 dataset are more able to illustrate this point, since the Coutrot2 records 4 persons having a meeting, and the audio plays a great role on human attention. In addition, Figure \ref{fig:4} depicts some visualized results. Expectedly, from the comparison results of the third row and the fourth row, we can clearly observe that the predicted saliency map is more accurate when the audio information is considered. Concretely, in the first video (the first two columns), when the audio information is ignored, the visual model locates the non-sounding person, which is no salient. This further demonstrate that the audio information can prompt the model to better locate the attention-grabbing sounding objects in the scene.

2) The devised audio-visual location part can locate the sounding objects effectively so as to enhance the performance of DSP methods. For the observation from TABLE \ref{tab:1}, we can find that when the devised audio-visual location is replaced with inner-product operation, the performance for DSP declines notably. For instance, the CC value is decreased from 0.6262 to 0.5151 on the AVAD dataset, when we adopt the AV Inner-Product method. This is because the devised audio-visual location part can be utilized to effectively locate the sounding objects, which more attract human attention.

3) The proposed multi-cues aggregation module can be capable to integrate the effects by multiple cues. From TABLE \ref{tab:1}, we can find the great differences between the results from Concatenate Fusion method and the proposed method. It is mainly because the Concatenate Fusion method can not effectively integrate the influences of multiple cues on saliency. In contrast, the proposed method can achieve this effectually, which indicates the proposed multi-cues aggregation module is able to integrate the effects by multiple cues.

4) The spatial attention contributes to model the higher-level semantic information for better DSP. Based on the comparison results between Proposed (w/o SA) and the proposed method, we can discover that the latter can get more superior performance on all benchmark datasets. Especially, the CC value on the ETMD dataset can be improved nearly 4\%, which indicates the importance of the spatial attention for capturing the higher-level semantic information.

5) The temporal attention lifts the performance of DSP methods by modeling motion information. The comparison results between Proposed (w/o TA) and the proposed method on TABLE \ref{tab:1} reveals the significance of the adopted temporal attention. It is because the temporal attention can capture the temporally moving information by exploiting the frame-wise similarity. The results verify the importance of the temporal attention for improving the performance of DSP methods.

\subsection{Comparison with State-of-the-arts}
To demonstrate the effectiveness of the proposed method, we compare the proposed method with 8 state-of-the-art DSP methods. These comparison methods are comprehensive, including 4 spatial methods and 5 spatio-temporal methods. 1) The spatial methods process each frame independently to generate the saliency map, while do not consider the temporal information among frames. The comparison spatial methods include DeepNet \cite{pan2016shallow}, DVA \cite{wang2017deep}, SAM \cite{cornia2018predicting}, and SalGAN \cite{pan2017salgan} methods. DeepNet \cite{pan2016shallow} addresses the DSP problem by utilizing convnets for regression in an end-to-end manner. DVA \cite{wang2017deep} learns multi-scale features for capturing hierarchical saliency information for DSP. SAM \cite{cornia2018predicting} predicts the saliency map by incorporating neural attentive mechanisms using a convolutional long short-term memory. SalGAN \cite{pan2017salgan} adopts the data-driven metric for DSP by training with an adversarial loss function.
2) The spatio-temporal methods generate the saliency map by capturing the spatial information within each frame and the temporal information among frames simultaneously. The comparison spatial methods include ACLNet \cite{wang2019revisiting}, DeepVS \cite{jiang2018deepvs}, TASED \cite{min2019tased}, and STAViS \cite{tsiami2020stavis} methods. ACLNet \cite{wang2019revisiting} employs the CNN-LSTM architecture for DSP with a supervised attention mechanism. DeepVS \cite{jiang2018deepvs} develops an object-to-motion convolutional neural network for estimating the intra-frame saliency. TASED \cite{min2019tased} exploits 3D fully-convolutional network architecture to generate the saliency map of each frame by considering several past frames. STAViS \cite{tsiami2020stavis} combines both visual and auditory information for DSP in videos.

\begin{table*}[t]
\caption{The quantitative comparison on six benchmark datasets.}
\centering
\scriptsize
\setlength{\tabcolsep}{1.5mm}{
\begin{tabular}{c|ccccc|ccccc}
\hline
\multirow{2}{*}{Methods}&
\multicolumn{5}{c|}{AVAD}&
\multicolumn{5}{c}{Coutrot1} \\
\cline{2-11}
  &CC$\uparrow$	   &NSS$\uparrow$	   &AUC-J$\uparrow$   &sAUC$\uparrow$   &SIM$\uparrow$    &CC$\uparrow$	   &NSS$\uparrow$	   &AUC-J$\uparrow$   &sAUC$\uparrow$   &SIM$\uparrow$ \\
\hline
DeepNet \cite{pan2016shallow}	&0.3831 	&1.85 	&0.8690	&0.5616 	&0.2564	&0.3402	&1.41 &0.8248	&0.5597	&0.2732		 \\
DVA	\cite{wang2017deep}    &0.5247	    &3.00	&0.8887	&0.5820	    &0.3633	&0.4306	&2.07	&0.8531	&0.5783	&0.3324	 \\
SAM	\cite{cornia2018predicting}    &0.5279	    &2.99 	&0.9025	&0.5777	    &0.4244	&0.4329	&2.11	&0.8571	&0.5768	&0.3672	 \\
SalGAN \cite{pan2017salgan}	&0.4912     &2.55 	&0.8865	&0.5799 	&0.3608	&0.4161	&1.85	&0.8536	&0.5799	&0.3321	 \\
\hline
ACLNet \cite{wang2019revisiting}	&0.5809     &3.17	&0.9053	&0.5600	    &0.4463	&0.4253	&1.92	&0.8502	&0.5429	&0.3612 \\
DeepVS \cite{jiang2018deepvs}	&0.5281     &3.01	&0.8968	&0.5858	    &0.3914	&0.3595	&1.77	&0.8306	&0.5617	&0.3174  \\
TASED \cite{min2019tased}	&0.6006     &3.16	&0.9146	&0.5898	    &0.4395	&0.4799	&2.18	&0.8676	&0.5808	&0.3884\\
STAViS \cite{tsiami2020stavis}	&0.6086     &3.18	&0.9196	&0.5936	    &0.4578	&0.4722	&2.11	&0.8686	&0.5847	&0.3935 \\
\bf{Proposed}   &\bf{0.6262} &\bf{3.57}	&\bf{0.9251}	&\bf{0.6203}	    &\bf{0.4820}	&\bf{0.4985}	&\bf{2.44}	&\bf{0.8798}	&\bf{0.6042}	&\bf{0.4154}\\
\hline\hline
\multirow{2}{*}{Methods}&
\multicolumn{5}{c}{Coutrot2}&
\multicolumn{5}{c}{DIEM}\\
\cline{2-11}
   &CC$\uparrow$	   &NSS$\uparrow$	   &AUC-J$\uparrow$   &sAUC$\uparrow$   &SIM$\uparrow$    &CC$\uparrow$	   &NSS$\uparrow$	   &AUC-J$\uparrow$   &sAUC$\uparrow$   &SIM$\uparrow$ \\
\hline
DeepNet \cite{pan2016shallow}	&0.3012	&1.82	&0.8966	&0.6000	&0.2019	&0.4075 	&1.52 	&0.8321	&0.6227 	&0.3183 \\
DVA	\cite{wang2017deep}    &0.4634	&3.45	&0.9328	&0.6324	&0.2742	&0.4779	    &1.97	&0.8547	&0.6410	    &0.3785	 \\
SAM	\cite{cornia2018predicting}    &0.4194	&3.02	&0.9320	&0.6152	&0.3041	&0.4930	    &2.05 	&0.8592	&0.6446	    &0.4261	 \\
SalGAN \cite{pan2017salgan}	&0.4398	&2.96	&0.9331	&0.6183	&0.2909	&0.4868     &1.89 	&0.8570	&0.6609 	&0.3931	 \\
\hline
ACLNet \cite{wang2019revisiting}	&0.4485	&3.16	&0.9267	&0.5943	&0.3229	&0.5229     &2.02	&0.8690	&0.6221	    &0.4279 \\
DeepVS \cite{jiang2018deepvs}	&0.4494	&3.79	&0.9255	&0.6469	&0.2590		&0.4523     &1.86	&0.8406	&0.6256	    &0.3923 \\
TASED \cite{min2019tased}    &0.4375	&3.17	&0.9216	&0.6118	&0.3142	&0.5579     &2.16	&0.8812	&0.6579	    &0.4615	\\
STAViS \cite{tsiami2020stavis}	&0.7349	&5.28 &{\bf 0.9581} &0.7106	&0.5111	&0.5795   &2.26	&0.8838	&0.6741	    &0.4824 \\
\bf{Proposed}   &\bf{0.7481}	&\bf{5.45}	&0.9537	&\bf{0.7294}	&\bf{0.5266}	&\bf{0.5924}     &\bf{2.33}	&\bf{0.8941}	&\bf{0.6982}    &\bf{0.4917}\\
\hline

\hline\hline
\multirow{2}{*}{Methods}&
\multicolumn{5}{c|}{ETMD}&
\multicolumn{5}{c}{SumMe} \\
\cline{2-11}
   &CC$\uparrow$	   &NSS$\uparrow$	   &AUC-J$\uparrow$   &sAUC$\uparrow$   &SIM$\uparrow$    &CC$\uparrow$	   &NSS$\uparrow$	   &AUC-J$\uparrow$   &sAUC$\uparrow$   &SIM$\uparrow$ \\
\hline
DeepNet \cite{pan2016shallow} 	&0.3879	&1.90	&0.8897	&0.6992	&0.2253	&0.3320	&1.55	&0.8488	&0.6451	&0.2274 \\
DVA	\cite{wang2017deep}    &0.4965	&2.72	&0.9039	&0.7288	&0.3165	&0.3983	&2.14	&0.8681	&0.6686	&0.2811 \\
SAM	\cite{cornia2018predicting}   &0.5068	&2.78	&0.9073	&0.7310	&0.3790	&0.4041	&2.21	&0.8717	&0.6728	&0.3272 \\
SalGAN \cite{pan2017salgan}	&0.4765	&2.46	&0.9035	&\bf{0.7463}	&0.3117	&0.3978	&1.97	&0.8754	&\bf{0.6882}	&0.2897 \\
\hline
ACLNet \cite{wang2019revisiting}	&0.4771	&2.36	&0.9152	&0.6752	&0.3290	&0.3795	&1.79	&0.8687	&0.6092	&0.2965 \\
DeepVS \cite{jiang2018deepvs}	&0.4616	&2.48	&0.9041	&0.6861	&0.3495	&0.3172	&1.62	&0.8422	&0.6120	&0.2622 \\
TASED  \cite{min2019tased}	&0.5093	&2.63	&0.9164	&0.7117	&0.3660	&0.4288	&2.10	&0.8840	&0.6570	&0.3337 \\
STAViS \cite{tsiami2020stavis}	&\bf{0.5690}	&2.94	&0.9316	&0.7317	&0.4251	&0.4220	&2.04	&0.8883	&0.6562	&0.3373 \\
\bf{Proposed} &0.5664	&\bf{3.05}	&\bf{0.9351}	&0.7406	&\bf{0.4325}	&\bf{0.4392}	&\bf{2.25}	&\bf{0.8945}	&0.6712	&\bf{0.3428} \\
\hline
\end{tabular}}
\label{tab:2}%
\end{table*}

\begin{figure*}[t]
\begin{center}
\includegraphics[width=\linewidth]{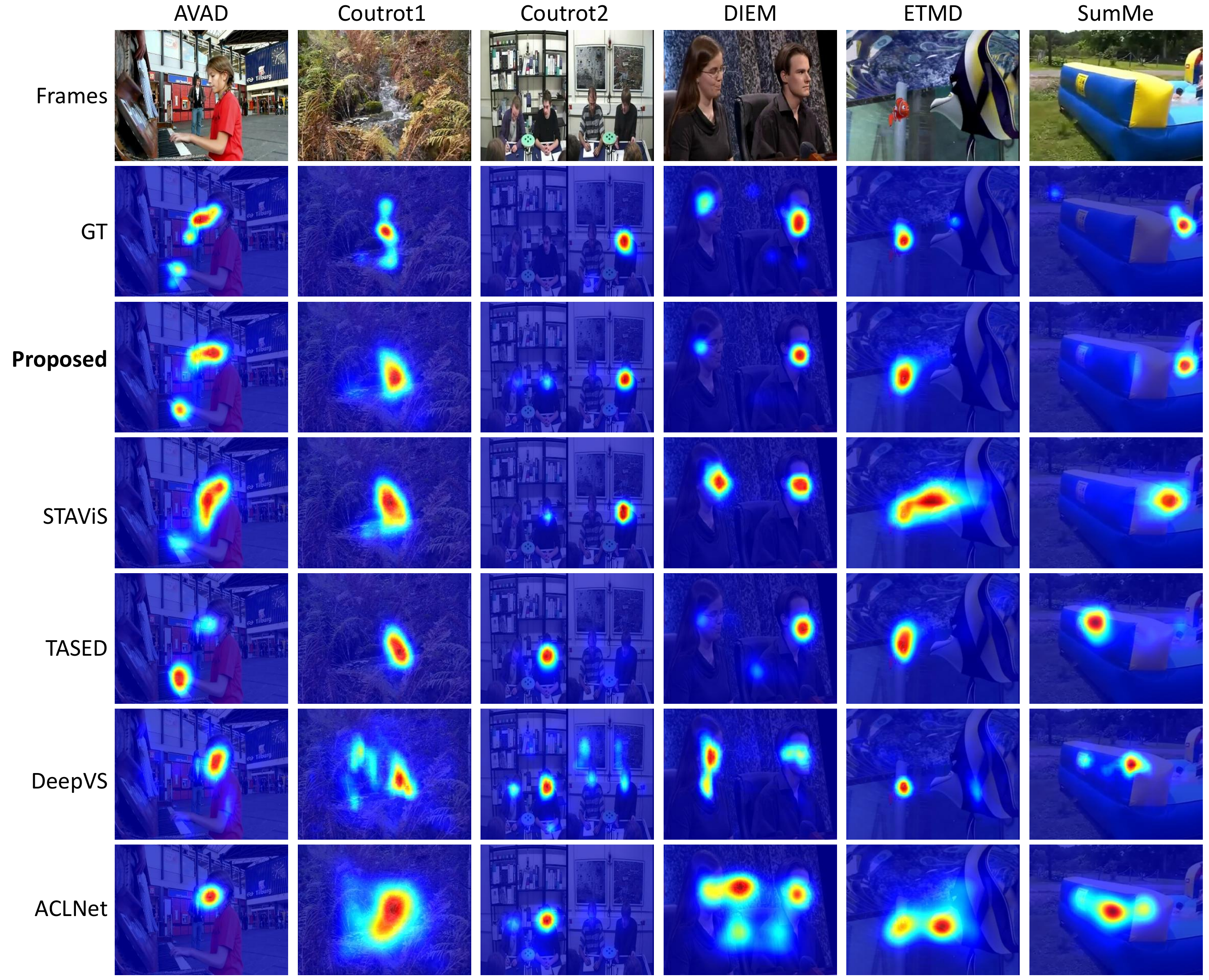}
\renewcommand{\figurename}{Fig.}
\vspace{-5 mm}
\end{center}
    \caption{\small{Qualitative results of different methods on six benchmark datasets. From top to bottom, each row represents the video frames, the corresponding groundtruth (GT), the predicted saliency map with the proposed method and other comparison methods.}}
\label{fig:5}
\end{figure*}

On the on hand, we compare the proposed method with 8 state-of-the-art DSP methods qualitatively on six benchmark datasets with audio-visual eye-tracking data, including AVAD \cite{min2016fixation}, Coutrot1 \cite{coutrot2016multimodal, coutrot2014saliency}, Coutrot2 \cite{coutrot2016multimodal, coutrot2014saliency}, DIEM \cite{mital2011clustering}, ETMD \cite{koutras2015perceptually, tsiami2019behaviorally} and SumMe \cite{gygli2014creating, tsiami2019behaviorally} datasets. TABLE \ref{tab:2} reports the qualitative results. As can be observed evidently, the proposed method outperforms the comparison methods with respect to most evaluation metrics. Especially, the proposed method surpasses the spatial DSP methods, such as DeepNet \cite{pan2016shallow}, DVA \cite{wang2017deep}, SAM \cite{cornia2018predicting}, and SalGAN \cite{pan2017salgan}, by a substantial margin. The success attributes to the proposed method capturing the temporal information, which is quite important for DSP. Compared with the spatio-temporal DSP methods, {\it e.g.}, ACLNet \cite{wang2019revisiting}, DeepVS \cite{jiang2018deepvs}, TASED \cite{min2019tased}, and STAViS \cite{tsiami2020stavis}, the proposed method also exhibits better performance. It is mainly because the proposed method can adaptively integrate multiple cues, which are essential for DSP. More specifically, even though the STAViS method also adopts audio-visual information for DSP, the proposed method surpasses it on the prediction of saliency map. This because we integrate more information affecting DSP, design more effective audio-visual location part for locating the sounding objects, and propose multi-cues aggregation module to collaboratively integrate the influence of multiple cues on dynamic saliency prediction. In addition, the higher CC and SIM values show that the generated saliency maps by the proposed method are more similar to the human annotations, which further demonstrates the superiority of the proposed method.

On the other hand, the qualitative visual comparisons are also conducted, and the results are shown in Figure \ref{fig:5}. As can be seen, the proposed method achieves the optimal performance among the comparison methods. Concretely, as for the sample in first column, the proposed method can locate both the sounding object (the piano) and high-level semantic object (human face) at the same time. In contrast, the TASED method mainly focuses on the sounding object. The DeepVS and ACLNet methods only pay attention to the high-level semantic object. The reason is that the proposed method collaboratively integrates multiple cues for DSP. Compared with the visualized results of STAVis, the proposed method can predict the attention-grabbing objects in the scene more accurately. It indicates the importance of other cues except for audio cues for DSP, and the superiority of audio-visual location part for locating sounding objects and the multi-cues aggregation module for adaptively integrating multi-cues information.

\section{Conclusions}  \label{conclutions}
In this paper, we propose an audio-visual collaborative representation learning method for dynamic saliency prediction based on the fact that human attention is naturally influenced by audio-visual stimuli, including sounding, moving, center, high-level semantic objects. Specifically, the proposed method consists of three main parts: audio-visual encoding, audio-visual location, and collaborative integration parts. The audio-visual encoding part encodes the input data as audio semantic features and spatio-temporal visual features. The audio-visual location part locates the sounding object by exploiting the consistency between the audio and visual modalities in a sharing latent space. The collaborative integration part captures the moving, center, and high-level semantic information, and adaptively integrates them with the sounding information to generate the final saliency map. The experimental results on six benchmark datasets have demonstrated that: 1) the audio information contributes to the DSP task significantly; 2) the consistency between the audio and visual modalities in a sharing latent space enhances the performance of sounding object location; 3) the adaptively aggregation of multi-cues helps the DSP method achieve the superior performance.

\section*{Acknowledgements}
This work was supported in part by Natural Science Basic Research Program of Shaanxi under Grant 2022JQ-592, in part by the Special Construction Fund for Key Disciplines of Shaanxi Provincial Higher Education, and in part by the Key Research and Development Program of Shaanxi Province under Grant 2021GY-181.


\bibliography{mybibfile}

\begin{thebibliography}{10}
\expandafter\ifx\csname url\endcsname\relax
  \def\url#1{\texttt{#1}}\fi
\expandafter\ifx\csname urlprefix\endcsname\relax\def\urlprefix{URL }\fi
\expandafter\ifx\csname href\endcsname\relax
  \def\href#1#2{#2} \def\path#1{#1}\fi

\bibitem{lai2022weakly}
Q.~Lai, T.~Zhou, S.~Khan, H.~Sun, J.~Shen, L.~Shao, Weakly supervised visual
  saliency prediction, IEEE Transactions on Image Processing 31 (2022)
  3111--3124.

\bibitem{jian2021integrating}
M.~Jian, J.~Wang, H.~Yu, G.-G. Wang, Integrating object proposal with attention
  networks for video saliency detection, Information Sciences 576 (2021)
  819--830.

\bibitem{fang2022lc3net}
X.~Fang, J.~Zhu, X.~Shao, H.~Wang, Lc3net: Ladder context correlation
  complementary network for salient object detection, Knowledge-Based Systems
  242 (2022) 108372.

\bibitem{ji2022dmra}
W.~Ji, G.~Yan, J.~Li, Y.~Piao, S.~Yao, M.~Zhang, L.~Cheng, H.~Lu, Dmra:
  Depth-induced multi-scale recurrent attention network for rgb-d saliency
  detection, IEEE Transactions on Image Processing 31 (2022) 2321--2336.

\bibitem{Zhang2017Online}
P.~Zhang, T.~Zhuo, W.~Huang, K.~Chen, M.~Kankanhalli, Online object tracking
  based on {CNN} with spatial-temporal saliency guided sampling, Neurocomputing
  257 (2017) 115--127.

\bibitem{zhu2021saliency}
M.~Zhu, G.~Hou, X.~Chen, J.~Xie, H.~Lu, J.~Che, Saliency-guided transformer
  network combined with local embedding for no-reference image quality
  assessment, in: Proceedings of the IEEE/CVF International Conference on
  Computer Vision, 2021, pp. 1953--1962.

\bibitem{yang2020saliency}
S.~Yang, Saliency-based image contrast enhancement with reversible data hiding,
  in: IEEE International Conference on Acoustics, Speech and Signal Processing,
  2020, pp. 2847--2851.

\bibitem{zhu2018spatiotemporal}
S.~Zhu, Z.~Xu, Spatiotemporal visual saliency guided perceptual high efficiency
  video coding with neural network, Neurocomputing 275 (2018) 511--522.

\bibitem{itti1998model}
L.~Itti, C.~Koch, E.~Niebur, A model of saliency-based visual attention for
  rapid scene analysis, IEEE Transactions on Pattern Analysis and Machine
  Intelligence 20~(11) (1998) 1254--1259.

\bibitem{bruce2007attention}
N.~Bruce, J.~Tsotsos, Attention based on information maximization, Journal of
  Vision 7~(9) (2007) 950--950.

\bibitem{gao2009decision}
D.~Gao, N.~Vasconcelos, Decision-theoretic saliency: computational principles,
  biological plausibility, and implications for neurophysiology and
  psychophysics, Neural Computation 21~(1) (2009) 239--271.

\bibitem{hou2007saliency}
X.~Hou, L.~Zhang, Saliency detection: A spectral residual approach, in: 2007
  IEEE Conference on Computer Vision and Pattern Recognition, 2007, pp. 1--8.

\bibitem{wang2017deep}
W.~Wang, J.~Shen, Deep visual attention prediction, IEEE Transactions on Image
  Processing 27~(5) (2017) 2368--2378.

\bibitem{cong2018review}
R.~Cong, J.~Lei, H.~Fu, M.-M. Cheng, W.~Lin, Q.~Huang, Review of visual
  saliency detection with comprehensive information, IEEE Transactions on
  circuits and Systems for Video Technology 29~(10) (2018) 2941--2959.

\bibitem{ma2022multi}
T.~Ma, W.~Tian, Y.~Xie, Multi-level knowledge distillation for low-resolution
  object detection and facial expression recognition, Knowledge-Based Systems
  (2022) 108136.

\bibitem{cheng2022tsgb}
L.~Cheng, P.~Fang, Y.~Liang, L.~Zhang, C.~Shen, H.~Wang, Tsgb: Target-selective
  gradient backprop for probing cnn visual saliency, IEEE Transactions on Image
  Processing 31 (2022) 2529--2540.

\bibitem{wang2022hierarchical}
B.~Wang, X.~Hu, C.~Zhang, P.~Li, S.~Y. Philip, Hierarchical gan-tree and
  bi-directional capsules for multi-label image classification, Knowledge-Based
  Systems 238 (2022) 107882.

\bibitem{mopuri2018cnn}
K.~R. Mopuri, U.~Garg, R.~V. Babu, Cnn fixations: an unraveling approach to
  visualize the discriminative image regions, IEEE Transactions on Image
  Processing 28~(5) (2018) 2116--2125.

\bibitem{xu2018personalized}
Y.~Xu, S.~Gao, J.~Wu, N.~Li, J.~Yu, Personalized saliency and its prediction,
  IEEE Transactions on Pattern Analysis and Machine Intelligence 41~(12) (2018)
  2975--2989.

\bibitem{wang2021semantic}
G.~Wang, C.~Chen, D.-P. Fan, A.~Hao, H.~Qin, From semantic categories to
  fixations: A novel weakly-supervised visual-auditory saliency detection
  approach, in: Proceedings of the IEEE/CVF Conference on Computer Vision and
  Pattern Recognition, 2021, pp. 15119--15128.

\bibitem{zhang2020spatial}
K.~Zhang, Z.~Chen, S.~Liu, A spatial-temporal recurrent neural network for
  video saliency prediction, IEEE Transactions on Image Processing 30 (2020)
  572--587.

\bibitem{di2020Discriminative}
D.~Hu, R.~Qian, M.~Jiang, X.~Tan, S.~Wen, E.~Ding, W.~Lin, D.~Dou,
  Discriminative sounding objects localization via self-supervised audiovisual
  matching, Advances in Neural Information Processing Systems 33 (2020)
  10077--10087.

\bibitem{di2020cross}
D.~Hu, X.~Li, L.~Mou, P.~Jin, D.~Chen, L.~Jing, X.~Zhu, D.~Dou, Cross-task
  transfer for geotagged audiovisual aerial scene recognition, in: Proceedings
  of the European Conference on Computer Vision, 2020, pp. 68--84.

\bibitem{song2022multimodal}
Q.~Song, B.~Sun, S.~Li, Multimodal sparse transformer network for audio-visual
  speech recognition, IEEE Transactions on Neural Networks and Learning Systems
  (2022).

\bibitem{qian2020multiple}
R.~Qian, H.~D. Di~Hu, M.~Wu, N.~Xu, W.~Lin, Multiple sound sources localization
  from coarse to fine, in: Proceedings of the European Conference on Computer
  Vision, 2020, pp. 292--308.

\bibitem{aytar2016soundnet}
Y.~Aytar, C.~Vondrick, A.~Torralba, Soundnet: learning sound representations
  from unlabeled video, in: Advances in International Conference on Neural
  Information Processing Systems, 2016, pp. 892--900.

\bibitem{hara2018can}
K.~Hara, H.~Kataoka, Y.~Satoh, Can spatiotemporal 3d cnns retrace the history
  of 2d cnns and imagenet?, in: Proceedings of the IEEE conference on Computer
  Vision and Pattern Recognition, 2018, pp. 6546--6555.

\bibitem{liu2021visual}
N.~Liu, N.~Zhang, K.~Wan, L.~Shao, J.~Han, Visual saliency transformer, in:
  Proceedings of the IEEE/CVF International Conference on Computer Vision,
  2021, pp. 4722--4732.

\bibitem{wang2018attention}
W.~Wang, J.~Shen, H.~Ling, A deep network solution for attention and aesthetics
  aware photo cropping, IEEE Transactions on Pattern Analysis and Machine
  Intelligence 41~(7) (2018) 1531--1544.

\bibitem{zhang2021saliency}
Y.~Zhang, M.~Jiang, Q.~Zhao, Saliency prediction with external knowledge, in:
  Proceedings of the IEEE Winter Conference on Applications of Computer Vision,
  2021, pp. 484--493.

\bibitem{kroner2020contextual}
A.~Kroner, M.~Senden, K.~Driessens, R.~Goebel, Contextual encoder--decoder
  network for visual saliency prediction, Neural Networks 129 (2020) 261--270.

\bibitem{zhang2021uncertainty}
J.~Zhang, D.-P. Fan, Y.~Dai, S.~Anwar, F.~Saleh, S.~Aliakbarian, N.~Barnes,
  Uncertainty inspired rgb-d saliency detection, IEEE Transactions on Pattern
  Analysis and Machine Intelligence (2021).

\bibitem{judd2009learning}
T.~Judd, K.~Ehinger, F.~Durand, A.~Torralba, Learning to predict where humans
  look, in: 2009 IEEE 12th International Conference on Computer Vision, 2009,
  pp. 2106--2113.

\bibitem{perazzi2012saliency}
F.~Perazzi, P.~Kr{\"a}henb{\"u}hl, Y.~Pritch, A.~Hornung, Saliency filters:
  Contrast based filtering for salient region detection, in: 2012 IEEE
  Conference on Computer Vision and Pattern Recognition, 2012, pp. 733--740.

\bibitem{wang2012visual}
Q.~Wang, Y.~Yuan, P.~Yan, Visual saliency by selective contrast, IEEE
  Transactions on Circuits and Systems for Video Technology 23~(7) (2012)
  1150--1155.

\bibitem{Vig2014Large}
E.~Vig, M.~Dorr, D.~Cox, Large-scale optimization of hierarchical features for
  saliency prediction in natural images, in: IEEE Conference on Computer Vision
  and Pattern Recognition, 2014, pp. 2798--2805.

\bibitem{Jiang2015SALICON}
M.~Jiang, S.~Huang, J.~Duan, Q.~Zhao, {SALICON}: Saliency in context, in: IEEE
  Conference on Computer Vision and Pattern Recognition, 2015, pp. 1072--1080.

\bibitem{he2019understanding}
S.~He, H.~R. Tavakoli, A.~Borji, Y.~Mi, N.~Pugeault, Understanding and
  visualizing deep visual saliency models, in: Proceedings of the IEEE
  Conference on Computer Vision and Pattern Recognition, 2019, pp.
  10206--10215.

\bibitem{Wang2018Deep}
W.~Wang, J.~Shen, Deep visual attention prediction, IEEE Transactions on Image
  Processing 27~(5) (2018) 2368--2378.

\bibitem{Sss2017DeepFix}
K.~Sss, K.~Ayush, R.~V. Babu, {DeepFix}: A fully convolutional neural network
  for predicting human eye fixations., IEEE Transactions on Image Processing
  26~(9) (2017) 4446--4456.

\bibitem{Liu2018Learning}
N.~Liu, J.~Han, T.~Liu, X.~Li, Learning to predict eye fixations via
  multiresolution convolutional neural networks, IEEE Transactions on Neural
  Networks and Learning Systems 29~(2) (2018) 392--404.

\bibitem{yuan2019bio}
Y.~Yuan, H.~Ning, X.~Lu, Bio-inspired representation learning for visual
  attention prediction, IEEE Transactions on Cybernetics 51~(7) (2021)
  3562--3575.

\bibitem{Kummerer2017Understanding}
M.~Kummerer, T.~S.~A. Wallis, L.~A. Gatys, M.~Bethge, Understanding low- and
  high-level contributions to fixation prediction, in: IEEE International
  Conference on Computer Vision, 2017, pp. 4799--4808.

\bibitem{li2017multiview}
X.~Li, M.~Chen, F.~Nie, Q.~Wang, A multiview-based parameter free framework for
  group detection, in: Thirty-First AAAI Conference on Artificial Intelligence,
  2017.

\bibitem{liVideoDistillation2021}
X.~Li, B.~Zhao, Video distillation, SCIENCE CHINA Information Sciences 51~(5)
  (2021) 695--734.

\bibitem{jiang2018deepvs}
L.~Jiang, M.~Xu, T.~Liu, M.~Qiao, Z.~Wang, Deepvs: A deep learning based video
  saliency prediction approach, in: Proceedings of the European Conference on
  Computer Vision, 2018, pp. 602--617.

\bibitem{bak2017spatio}
C.~Bak, A.~Kocak, E.~Erdem, A.~Erdem, Spatio-temporal saliency networks for
  dynamic saliency prediction, IEEE Transactions on Multimedia 20~(7) (2017)
  1688--1698.

\bibitem{gorji2018going}
S.~Gorji, J.~J. Clark, Going from image to video saliency: Augmenting image
  salience with dynamic attentional push, in: Proceedings of the IEEE
  Conference on Computer Vision and Pattern Recognition, 2018, pp. 7501--7511.

\bibitem{sun2018sg}
M.~Sun, Z.~Zhou, Q.~Hu, Z.~Wang, J.~Jiang, Sg-fcn: A motion and memory-based
  deep learning model for video saliency detection, IEEE Transactions on
  Cybernetics 49~(8) (2018) 2900--2911.

\bibitem{lai2019video}
Q.~Lai, W.~Wang, H.~Sun, J.~Shen, Video saliency prediction using
  spatiotemporal residual attentive networks, IEEE Transactions on Image
  Processing 29 (2019) 1113--1126.

\bibitem{wu2020salsac}
X.~Wu, Z.~Wu, J.~Zhang, L.~Ju, S.~Wang, Salsac: a video saliency prediction
  model with shuffled attentions and correlation-based convlstm, in:
  Proceedings of the AAAI Conference on Artificial Intelligence, Vol.~34, 2020,
  pp. 12410--12417.

\bibitem{wang2020video}
Z.~Wang, Z.~Zhou, H.~Lu, Q.~Hu, J.~Jiang, Video saliency prediction via joint
  discrimination and local consistency, IEEE Transactions on Cybernetics
  (2020).

\bibitem{chen2021video}
J.~Chen, H.~Song, K.~Zhang, B.~Liu, Q.~Liu, Video saliency prediction using
  enhanced spatiotemporal alignment network, Pattern Recognition 109 (2021)
  107615.

\bibitem{min2016fixation}
X.~Min, G.~Zhai, K.~Gu, X.~Yang, Fixation prediction through multimodal
  analysis, ACM Transactions on Multimedia Computing, Communications, and
  Applications 13~(1) (2016) 1--23.

\bibitem{sidaty2017toward}
N.~Sidaty, M.-C. Larabi, A.~Saadane, Toward an audiovisual attention model for
  multimodal video content, Neurocomputing 259 (2017) 94--111.

\bibitem{min2020multimodal}
X.~Min, G.~Zhai, J.~Zhou, X.-P. Zhang, X.~Yang, X.~Guan, A multimodal saliency
  model for videos with high audio-visual correspondence, IEEE Transactions on
  Image Processing 29 (2020) 3805--3819.

\bibitem{zhu2021lavs}
D.~Zhu, D.~Zhao, X.~Min, T.~Han, Q.~Zhou, S.~Yu, Y.~Chen, G.~Zhai, X.~Yang,
  Lavs: A lightweight audio-visual saliency prediction model, in: 2021 IEEE
  International Conference on Multimedia and Expo (ICME), 2021, pp. 1--6.

\bibitem{hu2020discriminative}
D.~Hu, R.~Qian, M.~Jiang, X.~Tan, S.~Wen, E.~Ding, W.~Lin, D.~Dou,
  Discriminative sounding objects localization via self-supervised audiovisual
  matching, Advances in Neural Information Processing Systems 33 (2020).

\bibitem{Afouras20b}
T.~Afouras, A.~Owens, J.~S. Chung, A.~Zisserman, Self-supervised learning of
  audio-visual objects from video, in: European Conference on Computer Vision,
  2020, pp. 208--224.

\bibitem{tsiami2020stavis}
A.~Tsiami, P.~Koutras, P.~Maragos, Stavis: Spatio-temporal audiovisual saliency
  network, in: Proceedings of the IEEE Conference on Computer Vision and
  Pattern Recognition, 2020, pp. 4766--4776.

\bibitem{chen2021audiovisual}
J.~Chen, Q.~Li, H.~Ling, D.~Ren, P.~Duan, Audiovisual saliency prediction via
  deep learning, Neurocomputing 428 (2021) 248--258.

\bibitem{tavakoli2019dave}
H.~R. Tavakoli, A.~Borji, E.~Rahtu, J.~Kannala, Dave: A deep audio-visual
  embedding for dynamic saliency prediction, arXiv preprint arXiv:1905.10693
  (2019).

\bibitem{senocak2018learning}
A.~Senocak, T.-H. Oh, J.~Kim, M.-H. Yang, I.~S. Kweon, Learning to localize
  sound source in visual scenes, in: Proceedings of the IEEE Conference on
  Computer Vision and Pattern Recognition, 2018, pp. 4358--4366.

\bibitem{chen2020deep}
Y.~Chen, X.~Lu, S.~Wang, Deep cross-modal image--voice retrieval in remote
  sensing, IEEE Transactions on Geoscience and Remote Sensing 58~(10) (2020)
  7049--7061.

\bibitem{pang2020multi}
Y.~Pang, X.~Zhao, L.~Zhang, H.~Lu, Multi-scale interactive network for salient
  object detection, in: Proceedings of the IEEE Conference on Computer Vision
  and Pattern Recognition, 2020, pp. 9413--9422.

\bibitem{zhang2019hierarchical}
Y.~Zhang, Y.~Yuan, Y.~Feng, X.~Lu, Hierarchical and robust convolutional neural
  network for very high-resolution remote sensing object detection, IEEE
  Transactions on Geoscience and Remote Sensing 57~(8) (2019) 5535--5548.

\bibitem{komodakis2017paying}
N.~Komodakis, S.~Zagoruyko, Paying more attention to attention: improving the
  performance of convolutional neural networks via attention transfer, in:
  International Conference on Learning Representations, 2017.

\bibitem{woo2018cbam}
S.~Woo, J.~Park, J.-Y. Lee, I.~S. Kweon, Cbam: Convolutional block attention
  module, in: Proceedings of the European Conference on Computer Vision, 2018,
  pp. 3--19.

\bibitem{cornia2018predicting}
M.~Cornia, L.~Baraldi, G.~Serra, R.~Cucchiara, Predicting human eye fixations
  via an lstm-based saliency attentive model, IEEE Transactions on Image
  Processing 27~(10) (2018) 5142--5154.

\bibitem{droste2020unified}
R.~Droste, J.~Jiao, J.~A. Noble, Unified image and video saliency modeling, in:
  European Conference on Computer Vision, 2020, pp. 419--435.

\bibitem{coutrot2016multimodal}
A.~Coutrot, N.~Guyader, Multimodal saliency models for videos, in: From Human
  Attention to Computational Attention, 2016, pp. 291--304.

\bibitem{coutrot2014saliency}
A.~Coutrot, N.~Guyader, How saliency, faces, and sound influence gaze in
  dynamic social scenes, Journal of Vision 14~(8) (2014) 1--17.

\bibitem{mital2011clustering}
P.~K. Mital, T.~J. Smith, R.~L. Hill, J.~M. Henderson, Clustering of gaze
  during dynamic scene viewing is predicted by motion, Cognitive Computation
  3~(1) (2011) 5--24.

\bibitem{koutras2015perceptually}
P.~Koutras, P.~Maragos, A perceptually based spatio-temporal computational
  framework for visual saliency estimation, Signal Processing: Image
  Communication 38 (2015) 15--31.

\bibitem{tsiami2019behaviorally}
A.~Tsiami, P.~Koutras, A.~Katsamanis, A.~Vatakis, P.~Maragos, A behaviorally
  inspired fusion approach for computational audiovisual saliency modeling,
  Signal Processing: Image Communication 76 (2019) 186--200.

\bibitem{gygli2014creating}
M.~Gygli, H.~Grabner, H.~Riemenschneider, L.~Van~Gool, Creating summaries from
  user videos, in: European Conference on Computer Vision, 2014, pp. 505--520.

\bibitem{bylinskii2018different}
Z.~Bylinskii, T.~Judd, A.~Oliva, A.~Torralba, F.~Durand, What do different
  evaluation metrics tell us about saliency models?, IEEE Transactions on
  Pattern Analysis and Machine Intelligence 41~(3) (2018) 740--757.

\bibitem{pan2016shallow}
J.~Pan, E.~Sayrol, X.~Giro-i Nieto, K.~McGuinness, N.~E. O'Connor, Shallow and
  deep convolutional networks for saliency prediction, in: Proceedings of the
  IEEE Conference on Computer Vision and Pattern Recognition, 2016, pp.
  598--606.

\bibitem{pan2017salgan}
J.~Pan, C.~C. Ferrer, K.~McGuinness, N.~E. O'Connor, J.~Torres, E.~Sayrol,
  X.~Giro-i Nieto, Salgan: Visual saliency prediction with generative
  adversarial networks, arXiv preprint arXiv:1701.01081 (2017).

\bibitem{wang2019revisiting}
W.~Wang, J.~Shen, J.~Xie, M.-M. Cheng, H.~Ling, A.~Borji, Revisiting video
  saliency prediction in the deep learning era, IEEE Transactions on Pattern
  Analysis and Machine Intelligence 43~(1) (2019) 220--237.

\bibitem{min2019tased}
K.~Min, J.~J. Corso, Tased-net: Temporally-aggregating spatial encoder-decoder
  network for video saliency detection, in: Proceedings of the IEEE
  International Conference on Computer Vision, 2019, pp. 2394--2403.

\end{thebibliography}

\end{document}